\documentclass[runningheads]{llncs}

\usepackage[T1]{fontenc}
\def\doi#1{\href{https://doi.org/\detokenize{#1}}{\url{https://doi.org/\detokenize{#1}}}}
\usepackage{graphicx}
\usepackage{bm}

\usepackage{booktabs}

\usepackage{xcolor}
\usepackage{hyperref}
\usepackage{listings}
\usepackage{inconsolata} % Optional: Adds a nice monospaced font

\lstdefinestyle{mystyle}{
    language=Python,
    frame=single,
    numbers=right,
    numbersep=5pt,
    captionpos=b,
    showstringspaces=false,
    basicstyle=\ttfamily\footnotesize,  % Default monospaced font
}

\usepackage{booktabs}% http://ctan.org/pkg/booktabs

\usepackage{amsfonts}

\usepackage[numbers,square]{natbib}

\usepackage[binary-units=true,detect-weight=true]{siunitx}

\DeclareSIUnit{\nothing}{\relax}
\DeclareSIUnit{\instr}{i}

\usepackage{xspace}

\newcommand\solver[1]{\textsc{#1}\xspace}
\newcommand\vampire{\solver{Vampire}}

\renewcommand{\ttdefault}{cmtt} % Default monospace before frontmatter

\newcommand\shorten[1]{#1}
\newcommand\supershorten[1]{}

\begin{document}
% One Queue to Rule Them All? \\ Reinforcing Vampire's Clause Selection
% Reinforcement Learning Perspective of Clause Selection
\title{
% Towards Clause Selection Reinforcement
% Reinforcing Clause Selection
Efficient Neural Clause-Selection Reinforcement%\thanks{
% the project RICAIP no. 857306 under the EU-H2020 programme,
% and the Cost action CA20111 EuroProofNet.
%}
}
\titlerunning{Efficient Neural Clause-Selection Reinforcement}
% If the paper title is too long for the running head, you can set
% an abbreviated paper title here
%
\author{Martin Suda\orcidID{0000-0003-0989-5800}} % \inst{1}
\authorrunning{M. Suda}
% First names are abbreviated in the running head.
% If there are more than two authors, 'et al.' is used.
%
\institute{Czech Technical University in Prague, Czech Republic \\
\email{martin.suda@cvut.cz}}
\maketitle              % typeset the header of the contribution
\begin{abstract}
Clause selection is arguably the most important choice point in saturation-based theorem proving. 
Framing it as a reinforcement learning (RL) task
is a way to challenge the human-designed heuristics of 
state-of-the-art provers and to instead automatically 
evolve---just from prover experiences---their potentially optimal replacement.

% Fails to explicitly mention "stateless"

In this work, we present a neural network architecture for scoring clauses for clause selection 
that is powerful yet efficient to evaluate.
Following RL principles to make design decisions,
we integrate the network into the \vampire theorem prover and train it from successful proof attempts.
An experiment on the diverse TPTP benchmark finds the neurally guided prover improves
over a baseline strategy, from which it initially learns---in terms of the number of in-training-unseen problems
solved under a practically relevant, short CPU instruction limit---by \SI{20}{\percent}.

% trochu by se ale dalo rict, ze "Framing is a way to challenge" je tam i proto, ze to tim snazim lip pochopit

% In this work, we explore an RL setup TODO \ldots
% deliberately close to the human-designed heuristics (picking a set of simple clause features passed as inputs to a feed-forward neural network)
% to obtain a replacement that is general-purpose and competitive in real time evaluation.
% We couple our learning operator with the automatic theorem prover \vampire
% and study (TODO: "study" is too weak) its empirical properties on the diverse TPTP benchmark.

% THIS IS GSD - not here not now!
% There are inherent limits to how many problems can be solved relying on a single, universally good, clause selection strategy. %of certain kind.
% In the second part of the paper, 
% We show that a whole space of complementary yet relevant strategies can be evolved by the gradient descent process 
% used to train the guiding network. 

\end{abstract}

\renewcommand{\ttdefault}{zi4} % Inconsolata for the rest of the document

\section{Introduction}

Reinforcement learning (RL) \cite{SuttonBartoBookNew} is a machine learning (ML) paradigm %\footnote{Tem.} 
in which an agent learns to make sequential decisions by interacting with an environment and maximizing cumulative rewards.
The impressive successes of RL in board games \cite{Silver1140} % Silver_2016
or on the Atari benchmark \cite{DBLP:journals/corr/MnihKSGAWR13} % ,DBLP:conf/aaai/HesselMHSODHPAS18} 
motivate researchers to apply RL to their own domains of interest.
It is exciting to have an unsupervised method search for a decision strategy unbiased
by our human preconceptions, since this intuitively increases the chances of discovering
brand new approaches.

In automatic theorem proving (ATP), we have seen the Monte-Carlo tree search \cite{DBLP:conf/ecml/KocsisS06}
guide a connection tableaux prover \cite{DBLP:conf/nips/KaliszykUMO18,DBLP:conf/tableaux/ZomboriUO21} 
or, more recently, proximal policy optimization \cite{DBLP:journals/corr/SchulmanWDRK17} applied to 
dynamically pick clause selection queues \cite{DBLP:conf/cade/SchulzM16} in a saturation-based prover \cite{DBLP:conf/flairs/McKeownS23}.
Despite these achievements, substantial performance improvements over the state of the art have so far only been
reported for systems employing the more straightforward supervised learning methods \cite{DBLP:conf/itp/JakubuvU19,DBLP:conf/frocos/Suda21,DBLP:conf/lpar/ChvalovskyKPU23}. Moreover, the published results predominantly focus on problem sets 
with a common origin and encoding, such as the Mizar40 set \cite{DBLP:journals/jar/KaliszykU15a},
arguably more amenable to ML techniques than the canonical theorem proving benchmark 
used by the CASC \cite{Sut16} competition,
% the Thousands of Problems for Theorem Provers (TPTP) 
the TPTP library \cite{Sut24}.

In this work, we apply a policy gradient RL method \cite{DBLP:journals/ml/Williams92} to train a neural-network-based clause selection heuristic  for the ATP \vampire{} \cite{DBLP:conf/cav/KovacsV13}. As a highly optimized system, \vampire{} can generate and evaluate thousands to millions of clauses within seconds.
Not to impede the prover's high inference speed with the new guidance, we design a neural network architecture 
that is powerful yet efficient to evaluate (Sect.~\ref{sect:neural}). Analyzing clause selection through the lens of RL (Sect.~\ref{sect:clause_selection_reinforcement}),
we propose a new learning operator for iteratively improving the network's performance from successful proof attempts (Sect.\ref{sect:operator}). 
We adopt a rather counterintuitive idea in this context and ignore the concept of a state to
make our agent govern a single, standalone clause selection queue (Sect.~\ref{sect:implement}).

Even on the diverse TPTP benchmark, our new system is able to improve 
over \vampire{}'s default strategy, from which it initially learns, by \SI{20}{\percent}.
This is done in a sanitized setting (using a train/test split) and for practically relevant 
proving times of approximately \SI{10}{\second} per problem. Moreover, we find that the trained
neurally-guided \vampire{} solves more than 100 problems of rating 1.0, roughly half of which
have not been recorded as proven by any system to date. 
Improvements of this magnitude are exceptionally rare in the literature, particularly in terms of solving previously unrecorded problems.

After the experiments (Sect.~\ref{sec:expers}),
the paper reviews related work (Sect.~\ref{sec:related}) and ends with a conclusion with outlooks for future work (Sect.~\ref{sec:conclude}).

\section{Background}

% TODO: say something not too stupid here?

In this section, we review the basic concepts related to saturation-based theorem proving and reinforcement learning. % TODO: some more?

\subsection{Saturation-Based Theorem Proving}

% We assume the reader to be generally familiar with the concept of saturation in theorem proving. 

The most successful provers for first-order logic such as E \cite{DBLP:conf/cade/0001CV19}, 
iProver \cite{DBLP:conf/cade/DuarteK20}, SPASS \cite{DBLP:conf/cade/WeidenbachDFKSW09} 
or Vampire \cite{DBLP:conf/cav/KovacsV13}, are \emph{refutational}, based on \emph{saturation}.
The former means they aim for a proof by contradiction: 
Given an input problem $P$ consisting of a set of axioms $\mathcal{A}_P$ and a conjecture $G_P$,
they start by negating the conjecture and computing an equisatisfiable clause normal form $\mathit{CNF}(\mathcal{A}_P \cup \{\neg G_P\}) = C_P$
and then set out to show that this set of clauses $C_P$ is unsatisfiable.

During the saturation process which follows, the provers strive to compute a closure of these 
clauses with respect to a selected inference system $\mathcal{I}$. \supershorten{This means
that for every inference from $\mathcal{I}$ whose premises are among the input clauses or have been previously derived,
there is an obligation to also derive and add such inference's conclusion.
For satisfiable inputs, the closure may only be computable in the limit.
For an unsatisfiable set of clauses,
however, and a complete inference system $\mathcal{I}$,
a fair saturation strategy is bound to eventually derive the empty clause, which together with its
derivation ancestors constitutes the sought proof.} Saturation is most often 
implemented by a \emph{given-clause algorithm}, which keeps track of 
% the already performed and pending 
inferences by working with two sets of clauses: the \emph{active} set $\mathcal{A}$ (initially empty)
and the \emph{passive} set $\mathcal{P}$ (initialized with the input clauses). It maintains an invariant that every inference with 
all premises in $\mathcal{A}$ has already been performed.
Iteratively, a clause $C$ is \emph{selected} from $\mathcal{P}$, moved to $\mathcal{A}$ (and thus \emph{activated}), 
and all inferences with at least one premise being $C$ and others coming from $\mathcal{A}$ are computed 
and their conclusions are added to $\mathcal{P}$.

% given clause algorithm
% active / passive
% Discount ?

There are several variants of the given-clause algorithm \cite{DBLP:journals/jsc/RiazanovV03}.
They mostly differ in how they deal with simplifications and redundancy elimination,
a crucial topic for performance, but not immediately relevant for our exposition.

% Vubec tu nemluvim o empty clause / proof, ale so what, prijde pozdeji

\paragraph{Clause Selection:} Heuristics are employed to decide, in each iteration, which clause from $\mathcal{P}$
to select next for activation. An imagined perfect heuristic would only select clauses from the yet-to-be-discovered proof
(all other selections constitute, in retrospect, a wasted effort). Such an ideal, however,
must be intractable to compute as it would essentially eliminate search from the proving task.
In practice, it is thus important with clause selection to carefully trade between the quality of the decisions and 
the computational effort spent on making them.% \footnote{Just that I know.}

Provers typically implement the passive clause container as a collection of \emph{priority queues}, each queue representing $\mathcal{P}$ sorted 
by a distinct \emph{clause evaluation function} (CEF). The two most commonly used CEFs are \emph{age} and \emph{weight},
the first encoding a preference for older clauses and the second a preference for clauses with fewer symbols. 
The prover then alternates between the queues under a specified \emph{ratio}
(called the pick-given ratio in OTTER’s manual \cite{DBLP:journals/corr/cs-SC-0310056})
and selects the best clause from the current queue in each iteration.

\subsection{Reinforcement Learning}

Reinforcement Learning (RL) is an optimization framework
that enables an \emph{agent} to learn an optimal decision-making \emph{policy} 
through trial and error by interacting with an \emph{environment} 
and receiving feedback in the form of a \emph{reward}.
In what follows, we quickly summarize the main RL concepts; we refer the interested reader to a standard textbook \cite{SuttonBartoBookNew}
for a thorough formal treatment.

We focus here on a conceptualization of RL via finite Markov Decision Processes (MPDs),
in which time is modeled as passing in discrete steps. % Ignoring the Markov(ian) property, since who cares
In each step, the agent reflects on the current \emph{state} $s$ of the environment and correspondingly chooses one of possible \emph{actions} to take.
One time step later, in part as a consequence of the chosen action $a$,
the agent receives a numerical reward $r$ and finds itself in a new state $s'$. This transition (including the reward)
is in general stochastic, specified by a probability distribution $p(s',r\,|\,s,a)$.

A policy is a mapping from states to probabilities of selecting each of the available actions, denoted $\pi(a \,|\, s)$.
A \emph{value} $v(s)_\pi$ of a state $s$ under a policy $\pi$ is the expected \emph{return}, i.e., the reward accumulated (optionally under discounting) over a trajectory starting at $s$ and following $\pi$. RL optimizes a policy to maximize the expected return (from any state). Thus, 
aiming for a high reward in the distance can be more important than collecting mediocre rewards immediately.

% Do we need "episode", perhaps not...
% reward accumulated over an entire \emph{episode} (the sequence of states from a start state to a terminal state),

% Do we need details on Discounting? 

\section{Clause Selection Reinforcement} % a.k.a. "Analysis"

\label{sect:clause_selection_reinforcement}

Let us consider a standard clause selection heuristic like the age-weight alternation,
as if driven by an RL agent. Such an agent monitors the prover's state and chooses appropriate actions 
to reach the goal of deriving the empty clause, ideally in the smallest number of steps possible.
Our idea is to use this perspective to guide our decisions about the design of a new agent for the task, 
backed by a neural network and learned through reinforcement from proving experiences.

One of our aims is to make sure that the new design accommodates the old heuristic
as an attainable point in the space of possible solutions. This provides for a useful sanity check,
as well as a good promise for the potential to go beyond the state of the art, at least as long
as the computational overhead stays relatively low and there is anything new and relevant to learn from the data.

% Consider self-citing Elements 
% Elements \cite{elementsAITP22}, Two Operators \cite{twoOperatorsAITP24}

\paragraph{State.} 

The environment's states arise from the actual prover states via an abstraction
that forgets any information not relevant for the agent's decisions. This natural modeling step
carries along the mentioned computational tradeoff: the more information the agent receives 
the more precise its decisions can in principle be, however, at the same time, 
the more expensive may the deciding become.

We can split the information relevant for clause selection into two conceptual parts:
a \emph{static} part, the problem $P$ given as input, and an \emph{evolving} part, 
consisting of any information changing during the proving process and 
influencing the preferences for clause selection.
This second part allows the agent to have a plan: ``First select a clause like this, when done go and select a clause like that''.

Surprisingly, state-of-the-art provers, backed by decades of research and experimentation, 
mostly ignore the evolving part for clause selection.\footnote{Static state enters the picture when selecting proving strategies or strategy schedules.}
Except possibly for a few bits to remember which queue to select the next clause from,\footnote{Although, 
within the RL framework, deterministic queue alternation might best be seen as a poor man's implementation of a probabilistic choice.}
the state's evolution is ignored and each selection aims greedily at the best available clause.
This could be mostly for efficiency reasons; after all, if the selection preferences were to change from state to state, % TODO: have a CEF introduced
we could no longer cheaply keep the passive set represented as an ordered queue.
On a more abstract level,
it is not clear how to take an advantage of the evolving state of a saturation-based prover,
consisting---at its generality---of the content of the active and passive set, large and quickly growing, but otherwise rather amorphous sets of clauses.\footnote{
Perhaps in contrast to some other methods, a state in saturation-based proving is 
best understood as a meta-state, representing concurrent search for many possible proofs at once, in analogy 
to the state of the $A^*$ algorithm with many open nodes, representing all the currently considered promising paths to a goal.} 

Speculations aside, we take this observation as an indication that proof planning is not 
a viable approach to general-purpose saturation-based proving (in its contemporary form)
and bake the assumption of trivial, single-state---effectively \emph{stateless}---environment into our design.
% So we noticed stateless of sota and use it as a working assumption
In a nutshell, this means we will not allow the agent to change its opinion about a clause during the prover run.

\paragraph{Actions.} % The least difficult decision seems to be about actions. 
At each iteration of the saturation loop, clause selection picks one of the clauses in the passive set for activation.
We therefore equate the set of available actions (at a given moment) with the current content of the passive set.

% many many multi-armed bandits all with a single brain:

A slightly strange consequence of this decision---in combination with the assumption that the score of a clause (as computed by a neural network)
should not change from one moment to the next---is that this single score % $l_C$ 
must allow the clause to play the role
of a bad decision when currently also accompanied by more promising clauses on the passive set, while at the same time
being the score of the best choice when all other passive clauses look worse.
We will see, however, that this mild oddity is no hurdle to learning in practice.

\paragraph{Reward.} An essential facet of any RL environment is the reward.
Since it is a priori not clear during saturation which clauses will eventually constitute the discovered proof,
the ideal, most faithful-to-reality environment should intuitively only assign a non-zero reward to the final,
empty-clause-deriving step. However, an agent would only have a chance to reasonably learn from 
such a sparse reward through massive exploration (trial and error) and value bootstrapping.\footnote{Bootstrapping in RL means updating estimates 
based on other estimates instead of waiting for the full return. It represents a solution to the \emph{credit assignment problem},
the challenge of determining which actions were responsible for a received reward.}  
As we already decided not to work with states, such avenue is effectively ruled out.

We instead follow most other ML-based approaches to clause selection guidance \cite{DBLP:conf/cade/ChvalovskyJ0U19,DBLP:journals/pami/AbdelazizCMACIK23,DBLP:conf/flairs/McKeownS23} in this regard, learn from the successful proof attempts only, 
and assign a reward to actions retrospectively, based on which clauses end up in the discovered proof.
In more detail, at every step, each passive clause $C$ that is in fact a future proof clause gets 
a reward $r_C=1$ and the remaining, non-proof clauses receive $r_C=0$. As detailed later (cf.~Sect.~\ref{sect:operator} below),
we also subject these rewards to several forms of scaling, with the intuitive idea to give each
solved problem a fair share in influencing the update of the trained network.

% We scale down the reward by the number of learning steps with the intention 
% of giving each solved problem the same strength in influencing the learned heuristic,
% no matter how many steps it took to solve the problem (while aiming to learn from every step in the solution).

\subsection{RL-Inspired Learning Operator}

\label{sect:operator}

% Operator means "Give me prover a bunch of successful prover runs -> I will update your a new network to be even better at this"

Let a \emph{trace} of a successful proof attempt on problem $P_T$ be a tuple 
\[T = (P_T,\mathcal{C},\mathcal{C}^+,\{\mathcal{P}_i\}_{i\in I_T}),\]
where $\mathcal{C}$ is the set of all input % (i.e., $C_{P_T} \subseteq \mathcal{C}$)
and derived clauses, % (with sufficient information attached so that these clauses can be evaluated by a given neural network architecture),
$\mathcal{C}^+ \subseteq \mathcal{C}$ marks clauses that ended up in the found proof,
and the $\mathcal{P}_i \subseteq \mathcal{C}$ are the snapshots of the content of the passive set
at each iteration $i\in I_T$ of the saturation loop just before clause selection.
By a \emph{learning operator} for clause selection we mean a procedure that receives as input a set of traces $\mathcal{T}$ 
and a neural network $N_{\bm{\theta}}$, described by a vector of learnable parameters $\bm{\theta}$,
and updates these parameters to obtain $\bm{\theta}'$,
so that $N_{\bm{\theta}'}$ is now better suited for solving the problems that gave rise to $\mathcal{T}$
and ideally also generalizes well to solve other problems.

The learning operator we describe here is derived from the REINFORCE algorithm and the accompanying policy gradient theorem \cite{DBLP:journals/ml/Williams92}.
This algorithm is an ideal approach for directly optimizing a policy\footnote{Another family of approaches, the value-based methods, 
work by learning (to approximate) the state value function and thus rely on meaningful (distinct) states.} using gradient descent.
We avoid introducing the algorithm in its full generality (see \cite{SuttonBartoBookNew} for more details) and instead immediately describe how it is reflected in our learning operator.
We start by recalling a standard trick from deep RL and the key theorem.

% ONE MORE BRIDGE SENTENCE

\paragraph{Logits and Softmax.}

We assume our network $N_{\bm{\theta}}$ produces a score $N_{\bm{\theta}}(C) = l_C$, usually called the \emph{logit}, 
for each available action, i.e., for each clause $C$ from the passive set $\mathcal{P}$ in our case.
The logits are to be normalized via the softmax function to yield a probability distribution
\[\pi_{C,\bm{\theta}} = \text{softmax}_C\bigl(\{l_D\}_{D \in \mathcal{P}}\bigr) = \frac{e^{l_C}}{\sum_{D \in \mathcal{P}} e^{l_D}}.\footnote{As such, 
the logits are dimensionless and have only relative meaning: $l_C - l_D = d$ signifies 
that clause $C$ should be $e^d$-times more likely to get selected than clause $D$.}\]
This allows us to construe the clauses' scores, at any given moment, as a stochastic clause selection policy.
The stochastic aspect is an inherent part of the theory, but not a necessary component in an implementation (cf.~Sect.~\ref{sect:intergration}).

% https://ufal.mff.cuni.cz/~straka/courses/npfl139/2324/slides/?06#22

\paragraph{Policy Gradient.} Let $\alpha>0$ be a learning step parameter. The key theorem behind REINFORCE tells us that
in order to improve a policy in terms of the expected return,
we should update the network parameters as in 
\begin{equation} \label{eq:reinforce_update}
\bm{\theta} \leftarrow {\bm{\theta}} + \alpha r_C \nabla_{\bm{\theta}} \log \pi_{C,\bm{\theta}}
\end{equation}
% ChatGPT: This rule effectively increases the probability of actions that led to higher rewards and decreases those leading to lower rewards.
i.e., in the direction of the gradient $\nabla_{\bm{\theta}} \log \pi_{C,\bm{\theta}}$ and with a magnitude proportional to the reward $r_C$.
(REINFORCE actually uses return here, the reward accumulated over the whole trajectory starting from the current decision $C$,
but in our effectively stateless setup this simplifies to the immediate reward.)

\supershorten{
% In a longer version, it might be interesting to drop a line here about how close this actually is to multi-label classification.
}

% The rest are natural adaptations to our context and to the assumption of stateless environment:

\paragraph{Our Operator.} Let $\mathcal{T}$ be a set of traces and $N_{\bm{\theta}}$ our network. We treat each moment in time in each of the given traces
as an independent opportunity to improve the clause evaluation function that $N_{\bm{\theta}}$ represents.
This also includes the moments in which the prover selected a different clause than a future proof clause,
for which the reward is 0, and thus strictly following (\ref{eq:reinforce_update}) would lead to no nontrivial update.
We instead always learn from all proof clauses that are on the passive set and average their contributions to the gradient.\footnote{
This is one of the aspects in which we depart from the pure RL paradigm and are choosing a pragmatic approach instead.}

\supershorten{
% In a longer version, it might be interesting to mention the many many multi-armed bandits here.
}

For every $T = (P_T,\mathcal{C},\mathcal{C}^+,\{\mathcal{P}_i\}_{i\in I_T})  \in \mathcal{T}$ and every $i \in I_T$, we define the \emph{good passive clauses at step $i$} as
$\mathcal{P}^+_i = \mathcal{P}_i \cap \mathcal{C}^+$ and the corresponding gradient contribution as
\[\delta^T_i =  \mathrm{mean}_{C \in \mathcal{P}^+_i} \nabla_{\bm{\theta}} \log \pi_{C,\bm{\theta}}.\]
Additionally, we set $\delta^T = \mathrm{mean}_{i \in I_T} \delta^T_i $ and $\delta = \mathrm{mean}_{T \in \mathcal{T}} \delta^T$
as the contribution of a single trace and of the whole set, resp. In experiments containing more than one trace for a single problem
we insist that ``all problems are equal'' (not to allow the more represented ones to have more strength) and instead use:
\[\delta^P = \mathrm{mean}_{T \in \mathcal{T},P_T = P}  \delta^T \quad\text{ and }\quad \delta^\mathit{fair} = \mathrm{mean}_{P, \exists T \in \mathcal{T},P_T = P}  \delta^P. \]

\paragraph{Iteration.} We start from a randomly initialized network $N_{\bm{\theta}_0}$. We can use it to guide our prover and generate the 
first set of traces $\mathcal{T}_1$ to learn from, although we expect such guidance to be quite poor. A better option might be to collect traces from runs guided 
by an already tuned clause selection heuristic, such as the standard age-weight alternation. In any case, we then apply the operator repeatedly, 
building a sequence of trace sets and guiding networks $\mathcal{T}_1,N_{\bm{\theta}_1},\mathcal{T}_2 \ldots$,
where network $N_{\bm{\theta}_j}$ is used to generate the traces $\mathcal{T}_{j+1}$ for the subsequent improvement round, until an optimum performance is reached.

%  - we should keep iterating to eventually reach the optimum

\section{Neural Clause Evaluation}
 
 \label{sect:neural}
 
The neural network (NN) architecture proposed in this work is designed to be general-purpose and efficient to evaluate.
The first property mandates \emph{name invariance}: we do not allow the network to base its decisions on concrete 
symbol names as these may change meaning from one input problem to the next.\footnote{For instance, in the TPTP library which
we target for our experiments, each individual problem comes with its own symbol signature, and any name overlap between
two problems' signatures can be considered purely coincidental.} 

\paragraph{One-off GNN Invocation.}

Previous work has established Graph Neural Networks (GNNs)
as a good basis for name invariant neural formula representations % DBLP:conf/nips/WangTWD17
\cite{DBLP:conf/ecai/OlsakKU20,DBLP:conf/cade/JakubuvCOP0U20,DBLP:journals/pami/AbdelazizCMACIK23}. 
However, GNNs are relatively expensive to evaluate and intuitively work best when given a large formula (i.e., many clauses)
to evaluate at once, so that individual constituents (sub-formulas, terms, symbols) provide sufficient context for one another \cite{DBLP:conf/tableaux/ChvalovskyJOU21,DBLP:conf/ijcai/FokoueACIKLM023}. This becomes tricky when evaluating clauses for clause selection,
as the group of newly derived clauses, which need to be evaluated after each activation, varies in size and is typically relatively small.

We avoid these complications by running a GNN only once, at the beginning of the saturation process on the input problem's CNF, % TODO: have CNF explained 
and have the GNN prepare 
name-invariant vectorial representations (so called \emph{embeddings}) of the signature symbols 
and the input clauses, to be further utilized in subsequent processing.

\paragraph{Generalizing Age and Weight with RvNNs.}

The idea is to use these embeddings to seed computations of two (independent) Recursive Neural Networks (RvNN) \cite{DBLP:conf/ki/KuchlerG96}.
One RvNN starts from the input clause embeddings and unrolls along the clause derivation tree (as in \cite{DBLP:conf/cade/000121a}),
the other starts from the symbol embeddings and unrolls along the clause parse tree (as in \cite{DBLP:conf/cade/ChvalovskyJ0U19}).
In fact, it is advantageous to treat each of these trees as a directed acyclic graph (DAG), share the common subgraphs and 
cache the results for the already computed subgraphs.\footnote{In an ATP implementing perfect term sharing,
we maintain at most one fixed-size embedding per shared subterm. This shows that the overhead
of the network maintenance does not increase the prover's asymptotic time (or space) complexity.}
 
One can think of these two RvNNs as allowing the training process to generalize and improve upon
the two most commonly used standard clause evaluation functions, age and weight. 
Clause age is essentially the depth of the derivation tree, while weight equals the number of nodes
in the parse tree. As such, they could in principle be learned even without the initial embeddings
from the GNN. However, we hope that the network discovers much more powerful CEFs still. % TODO: have the prelim section introduce CEF?

\paragraph{Completing the Neural CEF.} To complete the tower of NN modules and obtain
a final score for a clause, we concatenate the embeddings from the two RvNNs with a vector of 12 easy-to-compute
simple clause features. These features include the standard age and weight, the number of
literals based on their polarity, the number of variable occurrences or the number of AVATAR \cite{DBLP:conf/cav/Voronkov14} splits a clause depends on.\footnote{See Appendix~\ref{app:simple_clause_features} for a complete list.}
The concatenated vector is then fed to a simple fully-connected NN with one hidden layer and a single clause score output.
% TODO: maybe just add here a forward reference sentence to the Vampire implementation section
% - and perhaps also to Fig.~\ref{fig:argDiag}, as a summary of the just said?

\subsection{Note on Efficient RvNNs}

\label{sect:efficientRvNNs}

The general rule for making NN computation fast is to group as many operations as possible into one high-level one, 
which can be vectorized (such as matrix multiplication) and executed via a single (optimized) library call.
This is often obvious to do with ``rectangular`` or otherwise regular inputs such as images, 
but can be tricky with DAGs traversed by our RvNNs.

In contrast to our previous work \cite{DBLP:conf/cade/000121a,DBLP:conf/frocos/Suda21}, 
we strive here to maximize such grouping by postponing node evaluations as much as possible 
and enqueueing the pending operations into layers based on their dependencies.
The essence of the idea is captured in Fig.~\ref{fig:gage_enqueue} on the example of the generalized-age (\texttt{gage}) RvNN.

\begin{figure}[t]
    \centering
    \begin{minipage}{0.97\linewidth}
\begin{lstlisting}[style=mystyle]
def gage_insert(cl_num:int, inf_rule:int, parents:list[int]):
  level = max(base_level, 1 + max(height[p] for p in parents))
  height[cl_num] = level
  index = level - base_level
  if len(todo_layers) == index:
    todo_layers.append([])
  todo_layers[index].append((cl_num, inf_rule, parents))
\end{lstlisting}
    \end{minipage}
    \caption{Python code for inserting clause \texttt{cl\_num} derived by inference rule \texttt{inf\_rule} from parents \texttt{parents}
     for later processing by the generalized-age (\texttt{gage}) RvNN.} 
    \label{fig:gage_enqueue}
\end{figure}

To understand Fig.~\ref{fig:gage_enqueue}, % in Fig.~\ref{fig:gage_enqueue}, 
let us first explain its context and how it is initialized. 
% Apart from the clause embeddings, which are stored separately and do not enter the picture here, 
The code maintains for each clause its \texttt{height},
which the input clauses have set to 0 and the derived clauses get assigned here on line 3.
% Height is the least level not smaller than \texttt{base\_level} that's greater than clause's parents' levels.
The main task of the shown function is to insert a new clause (along with the information 
of how it got derived) into the buffer \texttt{todo\_layers} at the lowest possible index,
so that clauses at the individual layers are independent and can be evaluated together,
provided the embeddings for the lower layers have already been computed.

Thus, when later evaluating the inserted clauses, this can be done in a layer-by-layer fashion,
in a single bulk operation per layer. 
Thereupon, \texttt{base\_level} is incremented by the number of the processed layers 
% (\texttt{len(todo\_layers)}) 
and the \texttt{todo\_layers} buffer itself is cleared.
This arrangement has the property that \texttt{height} of a clause does not 
need to change once set, and the code works both during inference, where 
many evaluations are triggered over time, and during training, where a maximal ``compression''
can be enjoyed and the \texttt{todo\_layers} buffer is filled only once.\shorten{\footnote{
% TODO: this should be updated if we go 30K
In our experiments, the median height of the clause derivation tree was \num{10}.
% sudamar2@dai-04:/nfs/sudamar2/lawa$ ./scatter_stat.py ~/mtpa-gnn/newSplit30k/loop1/stats.pt
% gage_height_avg: 20.310721471687266
% gage_height_median: 10
% gage_width_avg: 2935.5183960908307
% gage_height_max: 16384
% gweight_height_avg: 17.243460764587525
% gweight_height_median: 6
% gweight_width_avg: 4443.355561943087
% gweight_height_max: 65538
% taken over the problems of the fixed 15000 problems from 
% the training set and over the traces generated by default vampire in 10K % TODO: consider dropping completely not to confuse with the impression 
}} % (implicit reference to end-to-end)
% have a little experimental section (fwd reference on this); or just footnote some median here and reference an appendix

One additional condition must, however, be met for the bulk operations to be possible,
namely that there is only a single ``combine'' operation for our RvNN
and that each node to evaluate can be represented by a fixed-size vector. 
In the example of our generalized-age network, we achieve this by concatenating 
a trainable inference rule embedding with exactly two parent embedding slots.
The first slot is reserved for the embedding of the main premise and the second 
averages the embeddings of all the remaining premises (if present).

\subsection{Architecture Details}

\begin{figure}[t]
    \centering
    \includegraphics[trim = 120mm 160mm 260mm 100mm, clip,scale=0.3]{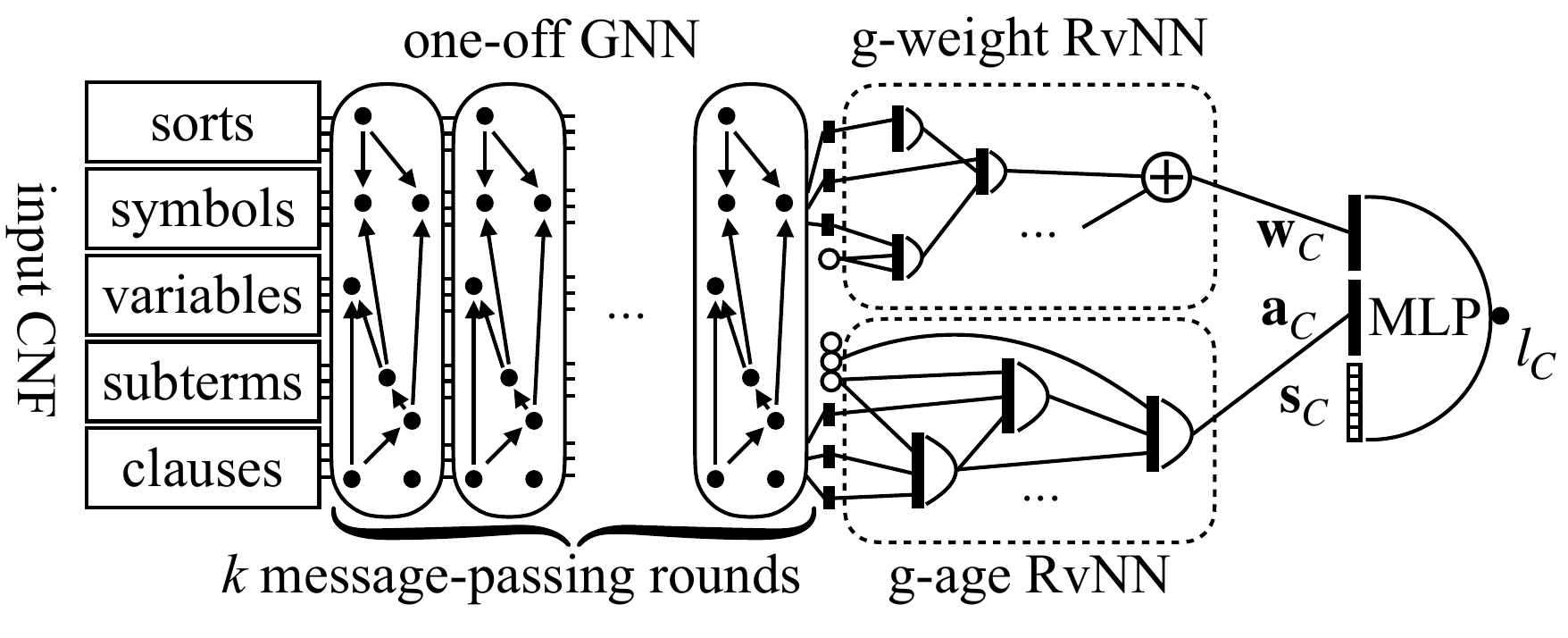} % Adjust width as needed
    \caption{Clause evaluation neural architecture. Information flows from left to right.}
    \label{fig:argDiag}
\end{figure}

A sketch of our architecture is shown in Fig.~\ref{fig:argDiag}. Following it from left to right,
we see how the input CNF is first turned into a graph consisting of five kinds of nodes 
(sort, symbol, variable, subterm, and clause nodes) and several kinds of edges. % (explained below).
The graph it is then processed by $k$ rounds of message-passing GNN, where $k$ is a hyperparameter,
producing embeddings for symbols and for the input clauses. The GNN is evaluated once, before saturation begins.

During saturation, the generalized-weight RvNN incrementally (on demand) produces embeddings of subterms, literals, and the derived clauses 
themselves, following their syntax structure. In addition to the symbol embeddings from the GNN
it also uses one trainable ``variable node'' embedding (denoted by an empty circle in Fig.~\ref{fig:argDiag}) to represent
any variable term (i.e., we deliberately conflate distinct variables and the RvNN produces the same embedding
vector for, e.g., both $r(X,Y)$ and $r(Y,X)$). The figure also highlights that a clause embedding here is obtained 
as a simple sum of its constituent literals' embeddings.

At the same time, the generalized-age RvNN uses the input-clause embeddings from the GNN and 
embeds the derived clauses by recursing along the clause derivation history. Fig.~\ref{fig:argDiag} shows that
each inner node combines the embeddings from the parents and from the used inference rule's trainable embedding 
(denoted by the empty circles).
Thus it is possible that, e.g., a clause $D$ derived from clauses $C_1$ and $C_2$ will obtain a different
embedding depending on whether it is derived by the subsumption resolution rule or by the binary resolution rule.

In the very right, Fig.~\ref{fig:argDiag} shows the final step of combining the embeddings from the two RvNNs with the
simple clause features mentioned earlier and passing them through a fully-connected neural block (MLP). Let us
spell out the mathematical details of this final step, because similar operations are also being performed
within the GNN and RvNN parts (and were just not detailed here). 

Let $n,m \in \mathbb{N}$ be the \emph{embedding size} and the \emph{expanded size} hyper-parameters, respectively.
For a clause $C$,
the final MLP receives the generalized weight and age embeddings $\mathbf{w}_C\in \mathbb{R}^n$ and $\mathbf{a}_C \in \mathbb{R}^n$, resp., 
and $C$'s simple features vector $\mathbf{s}_C \in \mathbb{R}^{12}$. The single hidden layer computes
\[\mathbf{h}_C = \text{ReLU}(\mathbf{W}_1\cdot [\mathbf{w}_C,\mathbf{a}_C,\mathbf{s}_C] + \mathbf{w}_2),\]
%i.e., 
an affine transformation with learnable tensors $\mathbf{W}_1 \in \mathbb{R}^{m \times (2n+12)}$ and $\mathbf{w}_2 \in \mathbb{R}^{m}$
followed by the standard activation function $\text{ReLU}(\mathbf{x}) = \max(\mathbf{0}, \mathbf{x})$.
And the final step is the linear $l_C = \mathbf{w}_3^T\cdot\mathbf{h}_C$, for a learnable vector $\mathbf{w}_3 \in \mathbb{R}^{m}$.\footnote{Further architecture details,
in particular the construction of the CNF graph and the computational details of the  GNN and the RvNNs have been
moved to Appendix~\ref{app:more_arch_details}.}

\section{Implementation}

\label{sect:implement}

We integrated the clause-selection-guiding NN into \vampire~4.9.\footnote{See \url{https://github.com/quickbeam123/deepire2.0-supplementary-materials}
for the details on how to obtain the system and reproduce the experiments.}
The prover's extension relies on the PyTorch~2.5 library \cite{NEURIPS2019_9015} and the {Torch\-Script} bridge for interfacing the neural model trained in Python from C++.
% Vampire (4.9) + pytorch; scripts are in python, jit script brings stuff back to C++ (cool stuff - what did I write in the Deepire papers?)
% -- fix the version of the experiments using a tag?
% available - footnote!
Although we initially aimed this way for zero code duplication between the Python training scripts and the invocation of the network from \vampire,
in the end, it was essential for good performance to duplicate the RvNN logic (cf.~Sect.~\ref{sect:efficientRvNNs}) in C++.\footnote{Relying on LibTorch, the C++ distribution of PyTorch.}
% subsection(Jit scripting)
% Jit script story (that's how we interface the NN written (mostly) in python), where it's also trained
% - that a hardcode implementation observation and sets up the scene
%
% - jit script story (perf and gather modes)
% -- great to have just one code for inference and training 
%  --- at least admit its in the end not the scripted code for the RvNNs
%
% libtorch magic could go here (how the same code is used for both training and inference; 
% although not quite - some things were simply too slow when done via python - this was major, so worth explaining
% (- also needed a non-system stack tree traversal to avoid stack overflows!)

\subsection{Single Queue Integration}

\label{sect:intergration}

When running under neural guidance, \vampire relies on a single queue ordered by the clause scores produced by the NN (the logits) to represent the passive set.
While properly sampling from the corresponding softmax distribution---as suggested by the theory (cf.~Sect.~\ref{sect:operator})---seemed too costly,
adding a small amount of noise to the scores once generated turned out to be a good way to diversify the search and solve additional problems on repeated runs.

% \cite{jang2017categorical}

Inspired by the Gumbel-max trick \cite{gumbel1954statistical,DBLP:conf/nips/MaddisonTM14}, we optionally add a Gumbel noise\footnote{Can be computed as $g = - \log(- \log(u))$ when sampling $u \sim \mathrm{Uniform}(0,1)$. } sample $g_C$ to the score of each clause (possibly scaled by a temperature parameter). The nice property this noise has
is that 
\[\arg\max\bigl( \{l_C + g_C\}_{C\in\mathcal{P}}\bigr) = \text{softmax}\bigl(\{l_C\}_{C \in \mathcal{P}}\bigr) \text{ (as distributions)},\]
so except for the fact that the Gumbel offsets $g_C$ are ``frozen'' (i.e., not drawn fresh for each clause selection round)
we achieve exactly the desired effect.

% subsection(Vamp1)
%
% - it should be clear we aim for a single CEF, the $\pi$ really
%
% Single queue + max / gumbel (no sampling)
% we also shuffle things to get more variance (ablation appendix only)
% -- shuffling to get more variability 

% ale tyhle body jsou spis implementacni detaily (a mohli by se resit v sekci "implementation in vampire")
% - jak bychom rádi samplovali, ale výpočetně je to pain (we had a sampling queue, but it's non triv)
% - Vlastně to moc není potřeba (max anyway wins)
% - Ale kdybychom přecijen, je tu gumbel trik (thanks, Jelle)
% - ale přestože max wins, docela se hodí mít víc dat (a k tomu ja randomizace dobra)

% GUBMEL:
% TODO: think of adding the discussion on efficient sampling and the gumbel-max trick there too!
% (https://timvieira.github.io/blog/post/2014/07/31/gumbel-max-trick/ + https://lips.cs.princeton.edu/the-gumbel-max-trick-for-discrete-distributions/ + https://arxiv.org/abs/1611.01144) 
%
% the theory here also talks about temperature

\shorten{
\subsection{Delayed Insertion Buffer}

\label{sect:delayed}

To group as many clauses as possible for a single bulk evaluation by the NN (cf.~Sect.~\ref{sect:efficientRvNNs})
while adhering to a principle that at the time of clause selection all passive clauses 
must already be assigned a score,\footnote{Related work \cite{DBLP:conf/cade/JakubuvCOP0U20} %,DBLP:conf/tableaux/ChvalovskyJOU21} 
explored % the option of 
postponing the evaluation of newly derived clauses until there is a certain number of them. 
This, however, may lead to the selection of a suboptimal clause while the current best clause is still waiting to be evaluated.}
we extended \vampire's passive clause container with a buffer of clauses waiting for evaluation 
before they can be inserted into the clause selection queue.
The buffer is important for the efficient maintenance of the passive set, 
because often a large fraction of newly derived clauses need not be evaluated and are instead forward-simplified (or deleted through subsumption) 
or clause splitting triggers changes on the passive set within the AVATAR architecture \cite{DBLP:conf/cav/Voronkov14},
all before it is necessary to know the clause scores for the next clause selection step.
}

% subsection(Vamp2)
% bulk evals and when to do them 
% - delayed evaluation buffer (is it also relevant outside LRS; even discount might pre-kill half of the new clauses)
% - but note that this is also quite low-level
%
% - delayed buffer (many new clauses don't get to passive; so don't eval them too early)

\subsection{Iterative Improvement Loop}

We use a collection of Python scripts to orchestrate 1) running \vampire to collect performance data,
2) rerunning the successful proof attempts in a mode that collects traces,\footnote{This incurs a performance penalty 
so unlike step 1) is run without a strict time limit. We made sure there is no non-determinism that would prevent reliable reproduction.} 
and 3) feeding these traces 
to our RL-inspired learning operator (cf.~Sect.~\ref{sect:operator}) to improve the guiding NN, all in a potentially infinite loop.
We use parallel processing to speed these operations as much as possible when run on a machine with multiple CPUs
(but do not make use of GPUs).
% multi-process training architecture (see 3.4 of our previous work~\cite{DBLP:conf/frocos/Suda21})

% validation traces
We make one noteworthy departure from RL practices in step 3). Instead of performing just one gradient descent step 
in each loop (as REINFORCE would), we always separate a random \SI{20}{\percent} of the available traces for validation purposes 
and have an inner loop of training rounds use the remaining \SI{80}{\percent} of the traces to repeat gradient descent until the validation loss\footnote{
Strictly speaking, Sect.~\ref{sect:operator} describes gradient \emph{ascent} with no explicitly loss. 
However, the role of a loss for gradient descent is in our case played by the expression $-\delta$ if understood ``before the gradient operator is applied.''} 
does not improve (for 5 successive rounds).
This is a standard \emph{early stopping} regularization criterion known from supervised learning, which in our case greatly accelerates convergence.\footnote{
The single-step and iterated %``superwised-like'' 
approach actually converge to the same fixed point \cite{DBLP:conf/nips/GhoshMR20}.}

\section{Experiments}

\label{sec:expers}

We use the TPTP library \cite{Sut24} v9.0.0 for our experiments. TPTP is very diverse, containing problems 
from many domains and of various encodings collected over several decades from many sources. 
% Does it belong here?
This poses a challenge for ML-based guiding methods as there are likely no immediate broad commonalities across the problems the learning could capitalize on.
Indeed, very few successes with improving ATPs via ML have been reported in the literature for this library to date.
% On the other hand, this makes the potential success all the more rewarding, because it suggests some general theorem proving knowledge is being picked up?

The first-order subset % (both CNF and FOF) 
of the library consists of \num{19477} problems and we randomly split those into \num{15000} training problems and the rest, left for independent testing.
For the main set of experiments we used an instruction limit\footnote{
Instruction limiting leads to more robust results than time limiting in situation (like ours) where many processes compete for the main memory (cf.~Appendix A of \cite{EasyChair:7719}).} of \SI{30000}{Mi} per proof attempt, 
% sudamar2@dai-02:~/mtpa-gnn-test$ /nfs/sudamar2/TPTP-v9.0.0/analyze_results_instr.py newSplittest_8668dai_otter_npcc-newSplit30kL20_i30000.pkl newSplitTest_8628_default_i30000.pkl
which amounts on average to \SI{10.4}{\second} of wall clock time on our servers.\footnote{Equipped with AMD EPYC 7513 (128 cores with \SI{2.6}{\giga\hertz}) and \SI{500}{\giga\byte} RAM.}
Unless stated otherwise, we fixed the NN parameters to: embedding size $n = 32$, expanded size $m=256$, and $k=8$ GNN rounds.
With these parameters, the neural model takes up a bit less than \SI{1.6}{\mega\byte} of disk space. % TODO: check for k=8!
% Fixing the architecture defaults (see appendix for more shit) and later exper sections for variations
% If we fix these and those HP's, we get a model file of XY  Mb

We use \vampire's \emph{default strategy} both as the source of initial training traces (relying on the default clause selection heuristic, 1:1 age-weight alternation),
as well as the basis for the neurally guided extension (with clause selection guided by the NN; cf.~Sect.~\ref{sect:intergration}).
The default strategy employs the AVATAR architecture \cite{DBLP:conf/cav/Voronkov14} and the limited resource strategy (LRS),
a version of the Otter saturation loop with a preemptive clause removal determined by an estimation of the speed of clause processing and the approaching time (instruction) limit \cite{DBLP:journals/jsc/RiazanovV03}.
Since delayed evaluation (cf.~Sect.~\ref{sect:delayed}) is in conflict with timely LRS estimations
and eager evaluation by the relatively expensive NN did not make LRS pay off anymore, we disabled LRS in the neurally guided version. 

% - related; LRS supported, but does not seem to pay off with a slow eval!
% -- lrs compatible (reproduce issue) : % that learning from it needs to solve the non-det issue 
% -- but not worth it with more expensive eval
% -- mini-exper; LRS does not pay off for neural (with a slow-ish classifier)
% (with a cheap one, it should help twice!)
% this all feels like a footnote material! 

\begin{figure}[t]
    \centering
    % sudamar2@dai-02:/nfs/sudamar2/lawa$ ./plotter.py ~/mtpa-gnn/newSplit30k-cumul-np5 ~/mtpa-gnn/newSplit30k  ~/mtpa-gnn/newSplit30k-noImit
    \includegraphics[scale=0.7]{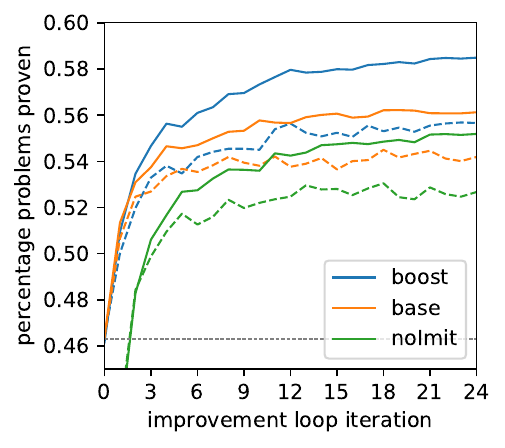}
    % sudamar2@dai-02:/nfs/sudamar2/lawa$ ./scatter.py ~/mtpa-gnn/newSplit30k/loop1/test_res.pt ~/mtpa-gnn/newSplit30k/loop20/test_res.pt
    \includegraphics[scale=0.7]{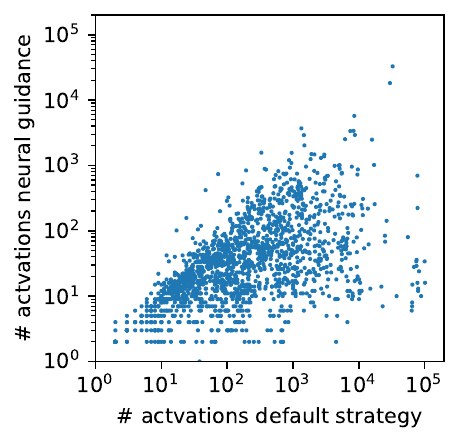}
    \caption{Left: performance progress of three selected training sessions (test performance dashed). Right: 
    scatter plot comparing the number of clause activations needed by the default and the neurally guided strategies, resp., on commonly solved test problems.}
    \label{fig:train_and_activations}
\end{figure}

\paragraph{Base experiment.}
The default strategy solves approximately \SI{46.3}{\percent} (\num{6940}) of the training problems.
Fig.~\ref{fig:train_and_activations} (left; \textsf{base}) shows the iterative improvement process
progressing from this baseline level (marked by the horizontal gray line)
to \SI{56.2}{\percent} training problems solved after 24 iterations.\footnote{The experiment took approximately 24 hours to complete,
when utilising 120 cores for the prover evaluation and trace collection and 60 cores for the training.} At the same time, the test performance
maxes with \SI{54.5}{\percent}, which means a \SI{17.8}{\percent} improvement over the baseline.
This shows that the trained model generalizes well from the training experience and enables \vampire to solve many new problems.

% Delegate to appendix 
% - LR and LR\_DECAY

% parallel perform (120 cores)
% parallel train (60 cores) mostly to keep the memory < 0.5T

\paragraph{Neural Guidance Cost and Quality.} 
% sudamar2@dai-04:/nfs/sudamar2/lawa$ ./scatter.py ~/mtpa-gnn/newSplit30k/loop1/test_res.pt ~/mtpa-gnn/newSplit30k/loop20/test_res.pt
% Network computation overhead analysis 2 - all failed neural runs (not just the wins of the default strategy)
% Not parsed 92
% Parsed but no gnn 113
% Avg nn_warmup 468.52394691286787
% Avg nn_bulks 9418.915432098765
% Avg nn_gnn 1741.0228395061729
Around one third of the \SI{30000}{Mi} allotted instructions (measured over unsuccessful runs that make it past parsing and preprocessing) 
is on average spent by the NN: \SI{470}{Mi} to load a model from the disc, \SI{1740}{Mi} by the GNN computation,
and \SI{9420}{Mi} on evaluating clauses during saturation. To compensate for this overhead,
the trained model offers a clause selection heuristic that is very precise. In Fig.~\ref{fig:train_and_activations} (right)
the neurally guided strategy needs on average \num{5.5}-times fewer clause activations than the default one and 
in only \SI{7.2}{\percent} of the cases it needs more activations to solve a given problem.
% sudamar2@dai-02:/nfs/sudamar2/lawa$ ./scatter.py ~/mtpa-gnn/newSplit30k/loop1/test_res.pt ~/mtpa-gnn/newSplit30k/loop20/test_res.pt
% geomeand vr1/vr2 act 5.540561518299103
% second more rate 0.07227488151658767

% Geoff and Jack: "reduce the number of steps used in successful proof searches as an indicator of intelligent search "

\supershorten{
After parsing and preprocessing the given problem, the GNN computation follows and must complete before saturation can start. 
There were 211 (\SI{4.7}{\percent}) large test problems for which the GNN computation did not finish in time. 
(Only 22 of these, however, got solved by the default strategy).
Among the remaining problems failed by the neurally guided prover because of the instruction limit,
an average of \SI{34.4}{\percent} of the given \SI{10000}{Mi} instructions were spent on evaluating clauses by the NN during saturation.
}

% TABULA RASA ASPECT of RL (= without imitation) -- can be postponed to experiments and related work (as we don't care in our story)
\paragraph{``Without Imitation.''} In addition to the base experiment, two other training sessions have their improvement progress shown in Fig.~\ref{fig:train_and_activations} (left). The \textsf{noImit} experiment starts the first loop iteration from a randomly initialized network (rather than the default strategy) to generate the first set of traces. While this is arguably more a property of the TPTP library\footnote{In that it contains enough problems so easy that
even an essentially random clause selection heuristics is able to solve them, 
yet sufficiently non-trivial ones that the corresponding solution traces can serve as a basis for subsequent improvement.} than of the training method, 
it is interesting to see that the baseline can be improved upon even without imitating the default clause selection heuristic.
However, the lower final performance suggests that the missing additional examples impair generalization.
% Without imitation; nice challenge, our approach copes, testament to TPTP (good curiculuning), but why bother if we can bootstrap from awr and thus rise higer? 

\paragraph{Boost.} The training session labeled in Fig.~\ref{fig:train_and_activations} (left) \textsf{boost} led to our current strongest guiding NN
trained from the \SI{30000}{Mi}-limited runs. It improves in test performance by \SI{20}{\percent} over the baseline.
The \textsf{boost} session relies on using several repeated runs of the prover to collect more training traces in each iteration,\footnote{
Instead of one run, we use 5 independently seeded runs with \vampire's input and internal shuffling \cite{DBLP:conf/cade/Suda22} enabled
and for the neurally guided runs we also add the Gumbel noise to the logits (cf.~Sect.~\ref{sect:intergration}) 
with temperatures $\tau \in \{0, 1/27, 1/9, 1/3, 1\}$.} and, additionally, assigns more weight in the training to (traces from) problems
solved only in recent loop iterations, while the weight of problems solved repeatedly in past successive loops is gradually decreased.\footnote{
Giving different weights to problems depending how useful they seem for the training appears to be a neat trick
we plan to explore more thoroughly in future work.}

\paragraph{Architecture Ablation Experiments.}

We conducted several experiments to establish the performance contributions of the individual architecture building blocks 
(the generalized-age RvNN, the generalized-weight RvNN, and the simple features) and the influence of the hyperparameters 
$n, m,$ and $k$. While the detailed results and several other technical details are deferred to Appendix~\ref{app:ablations}, % TODO: this will change in camera ready
we present here the main findings. 

If we disable just one of the three blocks, the performance drops only mildly and the remaining two blocks are able to compensate for the missing part.
We could speculate that this is (in part) because the simple features bring in the standard age and weight and can thus partially
act as proxies for their generalized versions. Leaving just one building block enabled, however, 
impairs performance substantially and the resulting NNs only barely improve over the baseline.

Reducing the embedding size ($n = 32$) or the expanded size ($m = 256$) 
from their respective base values leads to a noticeable decline in performance. 
Conversely, increasing these dimensions tends to improve results,
albeit at the cost of higher computational and memory requirements for the training phase.
Lastly, using $k = 8$ GNN rounds appears excessive in hindsight, 
as comparable performance can be achieved with only $k = 4$ rounds.

\paragraph{Solving Hard Problems.} To check whether our neural guidance can also help \vampire solve hard problems,
we ran one more training session, but this time with runs limited at \SI{100000}{Mi} ($\sim \SI{39.0}{\second}$) on the whole first-order TPTP.
% Spending 39.00142454248085 on average to burn 100K Mi.
% TODO: this was actually with a k=10 layer GNN!
Note that not using a train/test split is fine here, because none of the problems we now report were solved
by the seeding default strategy and so they had to be attained solely thanks to generalization. 
During the 30 improvement loops for which this experiment ran,\footnote{Taking \SI{12.4}{days} to complete, 
while using 64 cores for the prover evaluation and trace collection and 32 cores for training.} % (not to disturb other users of the server).}
it solved 130 TPTP problems of rating 1.0. % via sudamar2@dai-07:/nfs/sudamar2/lawa$ ./checkTheBang.py ~/mtpa-gnn/tptpOverfit100k3/
%%  When run reversed, 101 Rating 1.0 problems are covered by the final model

A problem of TPTP rating 1.0 has the property that no known system was able to solve it during the last rating evaluation \cite{Sut24}.
However, there are many rating 1.0 problems that had a rating smaller than 1.0 in the past. Out of our 130 problems,
we found 49 that were actually never solved even once in the past.\footnote{And 8 that had UNK status, 
so that it was not known until now whether they can be proven, or whether their conjecture is, in fact, false.}
This is a remarkable occurrence in the context of a single strategy improvement.
%
% ze ted je cele TPTP k dispozici na train
% ze tu generalizaci porad potrebujem, jinak by to nefungilo!
%
% ze to trvalo mnohem dyl (cf. jak to vlastne trva normalne?)
%
% ze jsou ruzne druhy TPTP rating 1.0; culminates with OPN

% sudamar2@dai-07:~/mtpa-gnn-test$ /nfs/sudamar2/TPTP-v9.0.0/analyze_results_instr.py newSplittest_8668dai_default_i1000000.pkl newSplittest_8668dai_default_nwc-5_i1000000.pkl newSplittest_8668dai_default_av-off_i1000000.pkl newSplittest_8668dai_otter_i1000000.pkl  newSplittest_8668dai_otter_npcc-newSplit30k-cumul-np5L28_i1000000.pkl  newSplittest_8668dai_otter_npcc-tptpOverfit100k3L26_i1000000.pkl
\begin{figure}[t]
    \centering
    \includegraphics[scale=0.7]{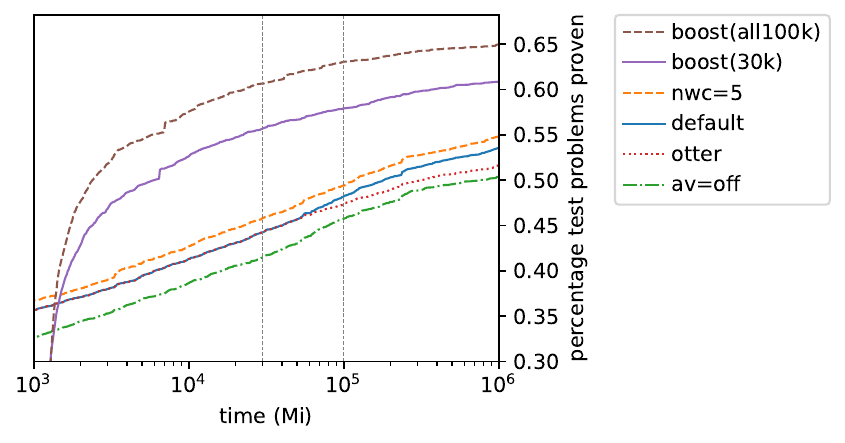}
    \caption[Test performance cactus plot]{Test performance cactus plot for \vampire's default strategy (and three variants) and two neurally guided strategies.
    % (Vertical line at \SI{10000}{Mi} denotes the limit used during the training.)
    (Vertical lines at \SI{30000}{Mi} and \SI{100000}{Mi}.)}
    \label{fig:mainCactus}
\end{figure}

As a comparison with the base line, we remark that there were only two TPTP rating 1.0 problems solved by the default strategy, % (LCL574+1, LCL575+1),
each solved only once out of the five randomly shuffled runs and both close to the limit of \SI{100000}{Mi}.

% TODO: zkus dopocitat do vyssich instrlimitu

\paragraph{Put Into Perspective.}

We close this section with a \emph{cactus plot} of Fig.~\ref{fig:mainCactus}, 
which shows the cumulative performance of several strategies as a function of time.\footnote{
    Another reference point, albeit not directly comparable, would be \vampire's CASC mode,
    a schedule of many strategies selected specifically so that they complement one another, 
    which, when run on 7 CPU cores for \SI{120}{\second}, covers \SI{68}{\percent} of the problems.}
First to notice is that the two neurally-guided strategies, \textsf{boost(all100K)} and \textsf{boost(30k)}, start solving any problems 
only after the \SI{1000}{Mi} mark, which is caused by the non-trivial start-up time of the guiding NN.
\textsf{boost(all100K)} is the strategy presented in the previous paragraph and, as noted, 
has been trained also on the testing split of our problems; as such it does not constitute a fair comparison and
is included only to provide a perspective.
The default strategy is shown in 4 variants: \textsf{default},
\textsf{otter} (which means turning off the LRS trick),
a conjecture-directed variant (\textsf{nwc=5})\footnote{NWC stands for ``non-goal weight coefficient'' and
means that any clause not derived from the conjecture will have its standard clause selection weight multiplied by 5.}
and a variant with AVATAR turned off. 

We would like draw the attention to the relative sizes of the jump in performance between
\textsf{av=off} and \textsf{default} and between \textsf{default} and \textsf{boost(30k)}.
When introduced, AVATAR came about as a major improvement in \vampire's performance,
now we can see a much larger jump thanks to our neural guidance.

\section{Related Work}

\label{sec:related}

The area of machine learning for theorem proving has been developing at a rapid pace in recent years \cite{DBLP:conf/birthday/BlaauwbroekCGJK24}.
Let us, therefore, focus here only on the most related approaches, those targeting the clause selection heuristic in saturation-based theorem proving.

% ENIGMA se popisuje jako supervised

% co z toho plyne:
% 1) pragmaticky to mame narocnejsi na data

% 2) trosku detailnejsi pohled:
% -- tim ze ENIGMA vidi jen selected ma sanci "opravit" tu heur, ze ktere se ucila
%    tezko se ale muze vymezit vuci klauzulim, ktere tam ani nedorazily, protoze je baseline run povazoval za blbe
% neprimo to vidime na 
%  A) enigma hlasi nejlepsi vysledky pri zapojeni v coopmodu
%  B) deepire (ktery je taky predstavitelem ENIGMA-style v tomto smyslu) hlasi nejlepsi vysledky 
%     pri zapojeni pres layered clause selection, kde se puvodni heuristika zachovava dokonce 
%     i na klauzulich vybranych modelem jako "good"

% My jsme prevzali RL jako naseho boha, ale ...

% enimga-style; list representatives, highlight differences (two operators will help)
Of these a prominent role is played by ENIGMA \cite{DBLP:conf/mkm/JakubuvU17,DBLP:conf/cade/ChvalovskyJ0U19,DBLP:conf/cade/JakubuvCOP0U20} and systems inspired by it \cite{DBLP:conf/lpar/LoosISK17,DBLP:conf/cade/000121a,DBLP:conf/frocos/Suda21}. Setting aside the ML technology used for now, % ,
we can single out a common \emph{ENIGMA-style learning operator} and compare it to the one proposed in this work.
While both approaches learn from successful prover runs and rely on proof clauses as their source of positive examples,
ENIGMA only uses the recorded selected clauses to form the negative background (as opposed to the typically much larger set of 
generated clauses used here), and, moreover, it puts these clauses onto one pile, abstracting away the flow of time 
(while the RL-inspired operator uses each clause selection moment in the trace as an independent situation to learn from).\footnote{
Chvalovsk{\'y} et al.~\cite{DBLP:conf/lpar/ChvalovskyKPU23} (Sect.~6.2) name this distinction \emph{classic} vs \emph{dynamic} data
and also prefer the latter for the training, as it is closer to the situation at inference time.}
ENIGMA can be seen to already assume a reasonably strong clause selection heuristic 
and mainly seeks to improve upon it through the integration of the learned guidance. This is reflected in the observation
that ENIGMA works best when combined with the baseline clause selection heuristic in some way (a.k.a. the ``coop'' mode), 
while the RL-inspired operator aims to provide independent guidance backed by a single clause selection queue.
Interestingly, while ENIGMA trains a binary classifier whereas the RL-inspired operator relies on the policy gradient 
theorem, the resulting formula for gradient descent is essentially the same (negative log likelihood).

% TODO: should the above be a table (in the journal)? but a back-to-back table could be also nice! (could happen in related work)
% maybe yes, because where else should One queue to rule them all? shine?

% TODO: promise experimental comparison against the ENIGMA-style operator for future work?

% iterative (looping) is no longer a thing to point out, right? (There was this planning paper where they had this very shy and precise analysis of when something might be maybe called "reinforced", when it learns from the successes its previous iteration help to generate.)

Among RL approaches, TRAIL \cite{DBLP:journals/pami/AbdelazizCMACIK23} also derives its loss from policy gradient,
% RL-based TRAIL - what's the difference? (they have state, similar "loss", probably too slow?)
but distinguishes itself by employing % multiplicative 
an attention mechanism to capture dependencies on the evolving state. 
% TRAIL is only evaluated on a small subset of the MIZAR benchmark, not on TPTP.
McKeown and Sutcliffe \cite{DBLP:conf/flairs/McKeownS23} use a value-based method, but only aim to learn a meta-heuristic:
% Jack - nice, but micekymouse results
an agent who picks which queue, out of the many provided by E prover \cite{DBLP:conf/cade/0001CV19}, to selected the next clause from.

% neural 
% - mirek -> Anonymous

Regarding neural architectures, the use of a GNN is not novel in our context \cite{DBLP:conf/ecai/OlsakKU20,DBLP:conf/cade/JakubuvCOP0U20,DBLP:journals/pami/AbdelazizCMACIK23}, %,DBLP:conf/lpar/ChvalovskyKPU23}, 
but the listed approaches apply it both to the input problem as well as to the derived clauses.
NIAGRA \cite{DBLP:conf/ijcai/FokoueACIKLM023}, a successor of TRAIL, similarly to our approach, uses a start-up invocation
% NIAGRA ma also nejakou "one-off GNN", ale ta story byla trochu jina (myslim, ze "one-off na celem vstupu; pak uz jen na jednotlivych klauzulich" ?)
of a GNN of the (often large) input problem to prepare symbol embeddings to later use cheaply during saturation. Embedding of formula 
syntax using RvNNs was historically the first use-case of the technology \cite{DBLP:conf/ki/KuchlerG96}, but more recently appeared, e.g., in ENIGMA-NG \cite{DBLP:conf/cade/ChvalovskyJ0U19}. An RvNN unrolling along clause derivation history comes from Deepire \cite{DBLP:conf/cade/000121a,DBLP:conf/frocos/Suda21}.

% TODO?
% - HER (also "without imitation") (also GNN; with Laplacian)

% TPTP vs Mizar?
% enigma-NG secret results (Thanks, Jan)
% our prelim res on Mizar?

% mention (with ackno) ENIGMA-anonymous' results on TPTP (5s) thanks to Jan

% mention in passing (footnote) our quick experiment with Mizar40 (it's a different ballpark!) // we have 15Mi run and a 50Mi followup (on dbai 79)

% HER nema TPTP (ale jeden z tech arxivovejch pre-HERu mel ten TPTP fragment - no equality there), co si pak vzali do NIAGRA ?

\section{Conclusion}

\label{sec:conclude}

We proposed a new neural architecture for clause selection guidance in an ATP and
a learning operator for its iterative improvement inspired by RL.
The architecture integrates a start-up GNN with two RvNNs, 
ensuring name invariance, efficiency in evaluation and strong performance 
thanks to access to both the clause's syntax as well as to its derivation history.
The learning operator assumes no evolving state as it suggests learning a single score pre clause, but 
draws experience independently from all the possibly many snapshots of the passive set along a successful proof attempt's trace.
The resulting neural clause selection heuristic is implemented as a single standalone clause selection queue. 
% Could still add a note about sampling?

On problems from the TPTP library, a particularly challenging benchmark for ML-based methods
due to its diverse nature, our new neural guidance improves the performance of \vampire{}'s default strategy by \SI{20}{\percent}
and, moreover, allows the prover to solve many problems not previously tackled by any known ATP.
To the best of our knowledge, this is a first published report of a substantial improvement 
by ML-based clause selection guidance on the TPTP benchmark.\footnote{In 2020 at CASC‑J10, ENIGMA Anonymous \cite{DBLP:conf/cade/JakubuvCOP0U20} improved 
% https://tptp.org/CASC/J10/WWWFiles/DivisionSummary1.html
over E prover, on which it is built, by 50 problems. The details of the success, however, remain unpublished.}

% TODO: RvNNs are needed should go to an appendix!

There are several interesting directions for future research. One of them is to experimentally compare
the relative strengths of the  ENIGMA-style and RL-inspired learning operators. Another is the question of transfer learning:
Would guidance trained on the TPTP enable \vampire{} to solve, e.g., more Mizar40 problems (or vice versa)?
Most importantly, we are curious about the potential of the new technology for improving 
whole sets of theorem proving strategies and the corresponding impact on building strong strategy schedules \cite{DBLP:conf/ijcar/BartekCS24}, as this is the ultimate measure of an improvement's impact on the ATP users.

\begin{credits}

\subsubsection{\ackname}
We thank the anonymous reviewers for their ideas for improvements.
We thank Jelle Piepenbrock and Micheal Rawson for suggesting the Gumbel-max trick for cheap softmax-like sampling.
The work was supported by the Czech Science Foundation project 24-12759S.

% Springer-mandated new weirdness.
\subsubsection{\discintname}
The authors have no competing interests to declare that are relevant to the content of this article.
\end{credits}

\bibliographystyle{splncs04}
\bibliography{main}

\begin{thebibliography}{10}
\providecommand{\url}[1]{\texttt{#1}}
\providecommand{\urlprefix}{URL }
\providecommand{\doi}[1]{https://doi.org/#1}

\bibitem{DBLP:journals/pami/AbdelazizCMACIK23}
Abdelaziz, I., Crouse, M., Makni, B., Austel, V., Cornelio, C., Ikbal, S.,
  Kapanipathi, P., Makondo, N., Srinivas, K., Witbrock, M., Fokoue, A.:
  Learning to guide a saturation-based theorem prover. {IEEE} Trans. Pattern
  Anal. Mach. Intell.  \textbf{45}(1),  738--751 (2023)

\bibitem{DBLP:journals/corr/BaKH16}
Ba, L.J., Kiros, J.R., Hinton, G.E.: Layer normalization. CoRR
  \textbf{abs/1607.06450} (2016)

\bibitem{DBLP:conf/ijcar/BartekCS24}
B{\'{a}}rtek, F., Chvalovsk{\'{y}}, K., Suda, M.: Regularization in
  spider-style strategy discovery and schedule construction. In: {IJCAR} 2024.
  LNCS, vol. 14739, pp. 194--213. Springer (2024)

\bibitem{DBLP:conf/birthday/BlaauwbroekCGJK24}
Blaauwbroek, L., Cerna, D.M., Gauthier, T., Jakub{\r u}v, J., Kaliszyk, C.,
  Suda, M., Urban, J.: Learning guided automated reasoning: {A} brief survey.
  In: Logics and Type Systems in Theory and Practice - Essays Dedicated to
  Herman Geuvers on The Occasion of His 60th Birthday. LNCS, vol. 14560, pp.
  54--83. Springer (2024)

\bibitem{DBLP:conf/tableaux/ChvalovskyJOU21}
Chvalovsk{\'{y}}, K., Jakub{\r u}v, J., Ol{\v s}{\'{a}}k, M., Urban, J.:
  Learning theorem proving components. In: {TABLEAUX} 2021. LNCS, vol. 12842,
  pp. 266--278. Springer (2021)

\bibitem{DBLP:conf/cade/ChvalovskyJ0U19}
Chvalovsk{\'{y}}, K., Jakub{\r u}v, J., Suda, M., Urban, J.: {ENIGMA-NG:}
  efficient neural and gradient-boosted inference guidance for {E}. In: {CADE}
  2019. LNCS, vol. 11716, pp. 197--215. Springer (2019)

\bibitem{DBLP:conf/lpar/ChvalovskyKPU23}
Chvalovsk{\'{y}}, K., Korovin, K., Piepenbrock, J., Urban, J.: Guiding an
  instantiation prover with graph neural networks. In: {LPAR} 2023. EPiC Series
  in Computing, vol.~94, pp. 112--123. EasyChair (2023)

\bibitem{DBLP:conf/cade/DuarteK20}
Duarte, A., Korovin, K.: Implementing superposition in {iProver}. In: {IJCAR}
  2020. LNCS, vol. 12167, pp. 388--397. Springer (2020)

\bibitem{DBLP:conf/ijcai/FokoueACIKLM023}
Fokoue, A., Abdelaziz, I., Crouse, M., Ikbal, S., Kishimoto, A., Lima, G.,
  Makondo, N., Marinescu, R.: An ensemble approach for automated theorem
  proving based on efficient name invariant graph neural representations. In:
  {IJCAI} 2023. pp. 3221--3229. ijcai.org (2023)

\bibitem{DBLP:conf/nips/GhoshMR20}
Ghosh, D., Machado, M.C., Roux, N.L.: An operator view of policy gradient
  methods. In: {NeurIPS} 2020 (2020)

\bibitem{gumbel1954statistical}
Gumbel, E.: Statistical Theory of Extreme Values and Some Practical
  Applications: A Series of Lectures. Applied mathematics series, U.S.
  Government Printing Office (1954)

\bibitem{DBLP:conf/nips/HamiltonYL17}
Hamilton, W.L., Ying, Z., Leskovec, J.: Inductive representation learning on
  large graphs. In: {NIPS} 2017. pp. 1024--1034 (2017)

\bibitem{DBLP:conf/cade/JakubuvCOP0U20}
Jakub{\r u}v, J., Chvalovsk{\'{y}}, K., Ol{\v s}{\'{a}}k, M., Piotrowski, B.,
  Suda, M., Urban, J.: {ENIGMA} {Anonymous}: Symbol-independent inference
  guiding machine. In: {IJCAR} 2020. LNCS, vol. 12167, pp. 448--463. Springer
  (2020)

\bibitem{DBLP:conf/mkm/JakubuvU17}
Jakub{\r u}v, J., Urban, J.: {ENIGMA:} efficient learning-based inference
  guiding machine. In: {CICM} 2017. LNCS, vol. 10383, pp. 292--302. Springer
  (2017)

\bibitem{DBLP:conf/itp/JakubuvU19}
Jakub{\r u}v, J., Urban, J.: Hammering {Mizar} by learning clause guidance
  (short paper). In: {ITP} 2019. LIPIcs, vol.~141, pp. 34:1--34:8. Schloss
  Dagstuhl - Leibniz-Zentrum f{\"{u}}r Informatik (2019)

\bibitem{DBLP:journals/jar/KaliszykU15a}
Kaliszyk, C., Urban, J.: Mizar 40 for mizar 40. J. Autom. Reason.
  \textbf{55}(3),  245--256 (2015)

\bibitem{DBLP:conf/nips/KaliszykUMO18}
Kaliszyk, C., Urban, J., Michalewski, H., Ol{\v s}{\'{a}}k, M.: Reinforcement
  learning of theorem proving. In: NeurIPS 2018. pp. 8836--8847 (2018)

\bibitem{DBLP:conf/ecml/KocsisS06}
Kocsis, L., Szepesv{\'{a}}ri, C.: Bandit based monte-carlo planning. In: {ECML}
  2006. LNCS, vol.~4212, pp. 282--293. Springer (2006)

\bibitem{DBLP:conf/cav/KovacsV13}
Kov{\'{a}}cs, L., Voronkov, A.: First-order theorem proving and {Vampire}. In:
  {CAV} 2013. LNCS, vol.~8044, pp. 1--35. Springer (2013)

\bibitem{DBLP:conf/ki/KuchlerG96}
K{\"{u}}chler, A., Goller, C.: Inductive learning in symbolic domains using
  structure-driven recurrent neural networks. In: {KI} 1996. LNCS, vol.~1137,
  pp. 183--197. Springer (1996)

\bibitem{DBLP:conf/lpar/LoosISK17}
Loos, S.M., Irving, G., Szegedy, C., Kaliszyk, C.: Deep network guided proof
  search. In: {LPAR} 2017. EPiC Series in Computing, vol.~46, pp. 85--105.
  EasyChair (2017)

\bibitem{DBLP:conf/nips/MaddisonTM14}
Maddison, C.J., Tarlow, D., Minka, T.: A* sampling. In: NIPS 2014. pp.
  3086--3094 (2014)

\bibitem{DBLP:journals/corr/cs-SC-0310056}
McCune, W.: {OTTER} 3.3 reference manual. CoRR  \textbf{cs.SC/0310056} (2003),
  \url{http://arxiv.org/abs/cs/0310056}

\bibitem{DBLP:conf/flairs/McKeownS23}
McKeown, J., Sutcliffe, G.: Reinforcement learning for guiding the {E} theorem
  prover. In: {FLAIRS} 2023. {AAAI} Press (2023)

\bibitem{DBLP:journals/corr/MnihKSGAWR13}
Mnih, V., Kavukcuoglu, K., Silver, D., Graves, A., Antonoglou, I., Wierstra,
  D., Riedmiller, M.A.: Playing {Atari} with deep reinforcement learning. CoRR
  \textbf{abs/1312.5602} (2013)

\bibitem{DBLP:conf/ecai/OlsakKU20}
Ol{\v s}{\'{a}}k, M., Kaliszyk, C., Urban, J.: Property invariant embedding for
  automated reasoning. In: {ECAI} 2020. Frontiers in Artificial Intelligence
  and Applications, vol.~325, pp. 1395--1402. {IOS} Press (2020)

\bibitem{NEURIPS2019_9015}
Paszke, A., Gross, S., Massa, F., Lerer, A., Bradbury, J., Chanan, G., Killeen,
  T., Lin, Z., Gimelshein, N., Antiga, L., Desmaison, A., Kopf, A., Yang, E.,
  DeVito, Z., Raison, M., Tejani, A., Chilamkurthy, S., Steiner, B., Fang, L.,
  Bai, J., Chintala, S.: Pytorch: An imperative style, high-performance deep
  learning library. In: Advances in Neural Information Processing Systems 32,
  pp. 8024--8035. Curran Associates, Inc. (2019)

\bibitem{DBLP:journals/jsc/RiazanovV03}
Riazanov, A., Voronkov, A.: Limited resource strategy in resolution theorem
  proving. J. Symb. Comput.  \textbf{36}(1-2),  101--115 (2003)

\bibitem{DBLP:journals/corr/SchulmanWDRK17}
Schulman, J., Wolski, F., Dhariwal, P., Radford, A., Klimov, O.: Proximal
  policy optimization algorithms. CoRR  \textbf{abs/1707.06347} (2017)

\bibitem{DBLP:conf/cade/0001CV19}
Schulz, S., Cruanes, S., Vukmirovic, P.: Faster, higher, stronger: {E} 2.3. In:
  {CADE} 2019. LNCS, vol. 11716, pp. 495--507. Springer (2019)

\bibitem{DBLP:conf/cade/SchulzM16}
Schulz, S., M{\"{o}}hrmann, M.: Performance of clause selection heuristics for
  saturation-based theorem proving. In: {IJCAR} 2016. LNCS, vol.~9706, pp.
  330--345. Springer (2016)

\bibitem{Silver1140}
Silver, D., Hubert, T., Schrittwieser, J., Antonoglou, I., Lai, M., Guez, A.,
  Lanctot, M., Sifre, L., Kumaran, D., Graepel, T., Lillicrap, T., Simonyan,
  K., Hassabis, D.: A general reinforcement learning algorithm that masters
  chess, shogi, and {Go} through self-play. Science  \textbf{362}(6419),
  1140--1144 (2018)

\bibitem{DBLP:conf/cade/000121a}
Suda, M.: Improving {ENIGMA}-style clause selection while learning from
  history. In: {CADE} 2021. LNCS, vol. 12699, pp. 543--561. Springer (2021)

\bibitem{DBLP:conf/frocos/Suda21}
Suda, M.: Vampire with a brain is a good {ITP} hammer. In: FroCoS 2021. LNCS,
  vol. 12941, pp. 192--209. Springer (2021)

\bibitem{EasyChair:7719}
Suda, M.: {Vampire} getting noisy: Will random bits help conquer chaos?
  EasyChair Preprint no. 7719 (2022),
  \url{https://easychair.org/publications/preprint/CSVF}

\bibitem{DBLP:conf/cade/Suda22}
Suda, M.: Vampire getting noisy: Will random bits help conquer chaos? In:
  {IJCAR} 2022. LNCS, vol. 13385, pp. 659--667. Springer (2022)

\bibitem{Sut16}
Sutcliffe, G.: {The CADE ATP System Competition -- CASC}. AI Magazine
  \textbf{37}(2),  99--101 (2016)

\bibitem{Sut24}
Sutcliffe, G.: {Stepping Stones in the TPTP World}. In: {IJCAR} 2024. pp.
  30--50. No. 14739 in LNAI (2024)

\bibitem{SuttonBartoBookNew}
Sutton, R.S., Barto, A.G.: Reinforcement Learning: An Introduction. A Bradford
  Book, Cambridge, MA, USA (2018)

\bibitem{DBLP:conf/cav/Voronkov14}
Voronkov, A.: {AVATAR:} the architecture for first-order theorem provers. In:
  {CAV} 2014. LNCS, vol.~8559, pp. 696--710. Springer (2014)

\bibitem{DBLP:conf/cade/WeidenbachDFKSW09}
Weidenbach, C., Dimova, D., Fietzke, A., Kumar, R., Suda, M., Wischnewski, P.:
  {SPASS} version 3.5. In: {CADE} 2009. LNCS, vol.~5663, pp. 140--145. Springer
  (2009)

\bibitem{DBLP:journals/ml/Williams92}
Williams, R.J.: Simple statistical gradient-following algorithms for
  connectionist reinforcement learning. Mach. Learn.  \textbf{8},  229--256
  (1992)

\bibitem{DBLP:conf/tableaux/ZomboriUO21}
Zombori, Z., Urban, J., Ol{\v s}{\'{a}}k, M.: The role of entropy in guiding a
  connection prover. In: {TABLEAUX} 2021. LNCS, vol. 12842, pp. 218--235.
  Springer (2021)

\end{thebibliography}

\newpage

\appendix

\section{List of the Used Simple Clause Features}

\label{app:simple_clause_features}

\begin{table}
\caption{Simple features of clause $C$. The domains $\mathbb{N}$ and $\mathbb{B}$ are only conceptual and get converted to $\mathbb{R}$ in the obvious way. }
\begin{tabular}{lcl}
short name & domain & description \\
\hline
age        & $\mathbb{N}$ & depth of $C$'s derivation tree \\
& &  (only counting generating inferences) \\
weight     & $\mathbb{N}$ & number of symbol occurrences in $C$ \\
% 3  & length     & $\mathbb{N}$ & number of literals in $C$ \\
posLen     & $\mathbb{N}$ & number of positive literals in $C$ \\
negLen     & $\mathbb{N}$ & number of negative literals in $C$ \\
justEq     & $\mathbb{B}$ & Are all literals equational? \\
justNeq    & $\mathbb{B}$ & Are all literals non-equational?\\
numVarOcc     & $\mathbb{N}$ & Number of variable occurrences in $C$ \\
numVarOccNorm & $\mathbb{R}^{[0,1]}$ & numVarOcc($C$) / weight($C$)           \\
fromGoal      & $\mathbb{B}$ & Does $C$ have a conjecture clause as an ancestor? \\
sineMaxed     & $\mathbb{B}$ & SInE fix point computation did not reach $C$ at all?  \\
sineLevelNorm & $\mathbb{R}^{[0,1]}$ & {\bf if} sineMaxed($C$) {\bf then} 1.0 \\
                         &  &  {\bf else} 0.5 $\times$ (sineLevel($C$) / maxSineLevelAssigned) \\
numSplits      & $\mathbb{N}$ & number of AVATAR assumptions that $C$ depends on \\
\end{tabular}
\end{table}

\section{Further Architecture Details}

\label{app:more_arch_details}

\paragraph{CNF Graph.} As mentioned, the graph corresponding to the input CNF 
has five kinds of nodes: sort, symbol, variable, subterm, and clause nodes.
It also has 8 distinct kinds of edges. (For the GNN message passing, though, opposite edges to each 
kind are added too, so there are in total 16 convolution kernels in each message passing round.)
Each node kind comes with a set of features to distinguish the individual nodes, if possible.

For an unsorted FOL problem, we introduce two sort nodes: \$i (default) and \$o (boolean),
where \$o is used as an output sort of predicate symbols. In a multi-sorted problem,
user-defined sorts are added; in a problem with arithmetic we also add \$int, \$rat, or \$real, as needed.
Sort nodes are represented by a feature vector of length 3, with Boolean features `isPlain', `isBoolean', and `isArithmetic'.

Symbols are both the predicate symbols and the function symbols. We use 10 features for symbols:
isEquality, isFunctionSymbol, isIntroduced (introduced by \vampire during preprocessing),
isSkolem (skolems are introduced, but there are other introduced symbols than skolems), isInterpretedNumber (such as 1 of sort \$int, 2.0 of sort \$real, etc.),
and arityGreaterThan$X$, where $X \in \{0,1,2,4,8\}$.

There is an edge (kind 1) from each symbol node to the symbol's output sort node.
There are also edges (kind 2) from symbols to larger symbols in the precedence.
These form a linear chain between the immediate neighbours,
but also regular jumps of length $2^i, i > 0$, so that in total not more than $n \log{n}$ edges are added. 
These ordering edges are added separately for the predicate symbol and function symbol nodes. 
We have not yet checked whether these edges meaningfully increase the expressive power of the GNN.

Variable, subterm and clause nodes are introduced together. They reflect the actual syntactic material of the input CNF,
but unlike the processing in the RvNNs, these are \emph{not shared} (allowing the latent meanings to be clause-local). 

Variable nodes are created for the variables of each clause separately. A clause $C = p(X,Y) \lor q(X)$ has two variables, 
$X$ and $Y$, so it causes two variable nodes to be added.
Variable nodes have a single feature, an ordinal number of that variable as it was encountered during the processing of its clause.
So if we add variable nodes $v_X$ and $v_Y$ for the sake of clause $C$, the node  $v_X$ will get feature $0$ and $v_Y$ feature $1$ or vice versa.
A variable node is connected to its clause's node by an edge (kind 3) and to its sort's node by another one (kind 4).

Subterm nodes are either literal nodes or proper subterm nodes, all the way down to terminal subterms: constant occurrences 
(function symbols of arity 0 applied to their 0 arguments) and variable occurrences. 
Each variable-occurrence node as a subterm node has an edge to a variable node (kind 5) of its variable.
Each non-variable-occurrence node has an edge (kind 6) to its functor's symbol node. Obviously, subterm nodes are connected by edges (kind 7)
reflecting the immediate subterm relation. Subterm nodes are represented by the following 10 features: 
literalSign (either $1$ or $-1$ for literal subterms, 0 for proper subterms), positionUnderParent (0 for literal subterms, linearly interpolating between 0.0 and 1.0 for proper subterms, e.g., in $p(a,b,c)$ this feature would be $0.0$ for $a$, $0.5$ for $b$ and $1.0$ for $c$),
numVariableOccurencesGreaterThan$X$ for $X \in \{0,2,4,8\}$ and weightGreaterThan$Y$ for  $Y \in \{1,4,16,64\}$.

Clause nodes are connected by an edge (kind 8) to their literals' subterm nodes. Each clause is represented by the following 10 features:
isDerivedFromGoal, isTheoryAxiom, hasSizeGreaterThan$X$ for $X \in \{1,2,4,8\}$ (i.e., the number of literals) and hasWeightGreaterThan$Y$ for $Y \in \{4,16,64,256\}$.

\paragraph{GNN Computation.}

For each of the five node kinds, we first pass the node feature matrix through a dedicated learnable affine transformation %  (\texttt{gnn\_node\_init[kind]})
to obtain an initial embeddings of size $n$ and apply the $\text{ReLU}$ non-linearity.

In each of the $k$ message-passing rounds that follow, GraphSAGE convolutions \cite{DBLP:conf/nips/HamiltonYL17}
are used to pass messages along each edge kind $\xi$. This means that new node embedding contributions $\mathbf{x}^\xi$ 
are computed from the old embeddings $\mathbf{x}$ via 
\[\mathbf{x}^\xi_i = \mathbf{W}^\xi_1 \mathbf{x}_i + \mathbf{W}^\xi_2 \cdot \mathrm{mean}_{j \in \mathcal{N}^\xi(i)} \mathbf{x}_j,\]
where $\mathcal{N}^\xi(i)$ are the nodes with an edge of kind $\xi$ leading to node $i$.

The overall new node embeddings $\mathbf{x}'$ are obtained by summing the contributions $\mathbf{x}^\xi$ over all edge kinds $\xi$
whose target node kind we are just discussing. For example, the embeddings for sorts
receive (and sum up) contributions from symbols (along edge kind 1) and variables (along edge kind 4).
Then the non-linearity $\text{ReLU}$ is applied to the new node embeddings $\mathbf{x}'$, 
they replace old node embeddings $\mathbf{x}$ and the process can be repeated. We stress that in each 
such message-passing round the convolution kernels (i.e., the matrices $\mathbf{W}^\xi_1$ and $\mathbf{W}^\xi_2$) are different
(we omitted the round index not to clutter the notation).\footnote{The intuition is that in each round 
the level of abstraction at which the data is internally represented may be shifting, from the rudimentary 
features we provide at the beginning, towards more high-level concepts that reflect the CNF more globally.}

As a final step, there is another set of learnable affine transformations applied 1) to the GNN clause embeddings, to promote them 
to the generalized-age RvNN initial clause embeddings, and 2) to the GNN symbol embeddings, to 
promote them to the generalized-weight RvNN symbol embeddings. 

% Left out:
% - that sorts also enter for Gweight (the sorts here help with symbols not part of input; e.g., numerals like 2 (=1+1)
% - that we don't need the kernels for some edges towards the end, as they would compute only information nobody reads

\paragraph{RvNN Computation.}

The generalized-age RvNN uses a matrix of trainable embeddings of the inference rules $\mathbf{W}^\mathit{inf} \in \mathbb{R}^{n \times r}$,
where $r = 205$ is the number of distinct inference rules recognized by \vampire.\footnote{While only a small subset of these rules is actually
utilized by the default strategy used in our experiments, it is convenient to allocate the full table 
and index the rules by their native id as obtained from the prover's source code.}
Its recursive step (the ``combine'' operation) is a single-hidden-layer MLP (similar to the final step already explained) and computes,
for a clause $C$ derived by inference rule $i$ from $n_p$ parents with already computed generalized-age embeddings $\mathbf{p}^1_C, \ldots, \mathbf{p}^{n_p}_C$,
\[\mathbf{h}^a_C = \text{ReLU}(\mathbf{W}^a_1\cdot [\mathbf{W}^\mathit{inf} \cdot e_i,\mathbf{p}^1_C,\mathrm{mean}_{k=2}^{n_p} \mathbf{p}^k_C] + \mathbf{w}^a_2),\]
where $\mathbf{W}^a_1 \in \mathbb{R}^{m \times 3n}, \mathbf{w}^a_2 \in \mathbb{R}^m$ and
the expression in the square bracket denotes a concatenation of 1) the $i$-th rule embedding, 2) the first parent embedding, and 3) the averaged remaining parent embeddings 
(as already mentioned at the end of Sect.~\ref{sect:efficientRvNNs}).
One additional affine transformation completes the generalized-age embedding of clause $C$ via
\[\mathbf{a}_C =  \text{LayerNorm}(\mathbf{W}^a_3\cdot \mathbf{h}^a_C + \mathbf{w}^a_4),\]
where $\mathbf{W}^a_3 \in \mathbb{R}^{n \times m}, \mathbf{w}^a_4 \in \mathbb{R}^n$
and $\text{LayerNorm}$ is the layer normalization step \cite{DBLP:journals/corr/BaKH16},
important for the stability of training involving many nested recursion steps. 

The generalized-weight RvNN recursive step is analogous in many respects.
There is a single variable subterm embedding $\mathbf{w}^\mathit{var} \in \mathbb{R}^{n}$ retrieved when an argument subterm is a variable.
We also use a single real number to encode the polarity $p$ of each subterm, similarly to how it is done in the GNN
(i.e., proper subterms have $p = 0$, positive literals $p=1$ and negative literals $p=-1$). % we don't care whether the functor is pred or func
So, assuming a term $t$ to embed
has functor $f$ with a symbol embedding $\mathbf{s}_f \in \mathbb{R}^n$,
polarity $p \in \mathbb{R}$, and $k$ already embedded argument subterms $s_1, \ldots, s_k$ (some of which may be variables) 
with respective embeddings $\mathbf{s}_1, \ldots, \mathbf{s}_k$, we first compute
\[\mathbf{h}^w_t =   \text{ReLU}(\mathbf{W}^w_1\cdot [\mathbf{s}_f,p,\mathbf{s}_1, \mathrm{mean}_{j=2}^{k} \mathbf{s}_j] + \mathbf{w}^w_2),\]
with $\mathbf{W}^w_1 \in \mathbb{R}^{m \times 3n+1}, \mathbf{w}^w_2 \in \mathbb{R}^m$, replacing $\mathbf{s}_1$ with $\mathbf{0} \in \mathbb{R}^n$ whenever $k=0$,
and follow it up with 
\[\mathbf{t} =\text{LayerNorm}(\mathbf{W}^w_3\cdot \mathbf{h}^w_t + \mathbf{w}^w_4),\]
with $\mathbf{W}^w_3 \in \mathbb{R}^{n \times m}, \mathbf{w}^w_4 \in \mathbb{R}^n$.
As mentioned, the generalized-weight embedding of a clause $C = L_1 \lor \ldots \lor L_k$
is computed simply as the sum
\[\mathbf{w}_C = \sum_{i=1}^{k}{\mathbf{l}_i},\]
with $\mathbf{l}_i$ standing for the subterms embeddings of the respective literals $L_i$.

% Here, so far, thinking only about the network evaluation, show make a remark that its meant to be end-to-end trainable? But training should better be explained elsewhere.

\section{Architecture and Training Setup Ablations}

\label{app:ablations}

In addition to the hyper-parameters mentioned in the main text, we used a base learning rate $\alpha = 0.0002$,
which decayed exponentially with each improvement iteration by a factor of $0.87055$, i.e., halving every five iterations.

Unlike in the main text, all experiments reported here used an instruction limit of \SI{10000}{Mi}
and, for the GNN, $k=10$ message passing rounds by default.

\paragraph{Adaptive Problem Weights.} In addition, all experiments reported here
applied the adaptive problem weighting scheme (mentioned in the main text as part of the \textsf{boost} session).
In more detail, the scheme works as follows. 

Given basic hyper-parameters $\mathit{maxStrength}=2.0$ and $\mathit{staleAfter}=5$ and a derived
parameter \[\mathit{base}=\mathit{maxStrength}^{\frac{1}{2*\mathit{staleAfter}}},\]  % math.pow(HP.CUM_MAX_STRENGTH,1/(2*HP.CUM_STALE_AFTER))
each solved problem $P$ maintains its score initialized to $\mathit{score}_P = 0$. 
The idea is that a relative weight of the problem's trace(s) in the overall loss expression
is $\mathit{base}^\mathit{score}_P$, i.e., starting at 1, and that $\mathit{score}_P$ is updated as follows.
Each iteration in which the problem is solved, its score is decreased by $1$, each time it is not solved,
its score is increased by $2$. However, if it is not solved for $\mathit{staleAfter}$ successive iterations,
the problem is declared ``stale'' and is not learned from anymore. The overall intuition is that easy problems
should contribute less than those the current version of the guidance struggles with.

In a particular improvement session (referred to as \textsf{base10k} below), the seeding default strategy solved \num{6529} problems,
which gave rise to \num{5560} traces to learn from (really easy problems are solved without requiring any clause selection steps
and cannot be learned from). After 30 iterations, \num{4999} problems ended up with the lowest possible score $-29$ and more than a thousand additional problems were assigned score from $\{-28,-27,-26\}$. A maximal final score was 34, for a problem that necessarily alternated between being solved and not being solved through consecutive iterations several times.

\begin{figure}
    \centering
    \includegraphics[scale=0.7]{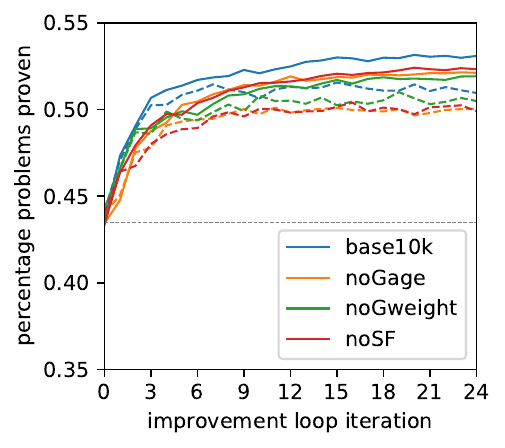}
    \includegraphics[scale=0.7]{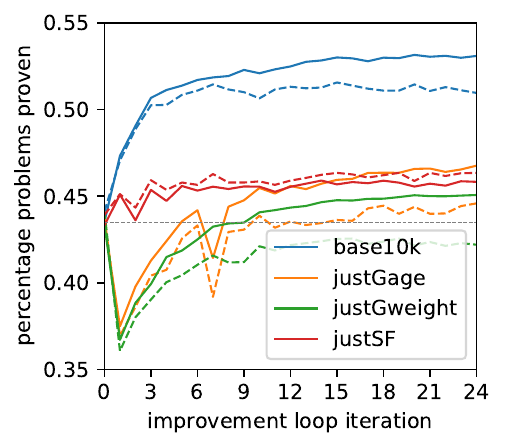}
    \caption{Performance progress
    for variants of our NN architecture with a single building block disabled (left) and a single block enabled (right),
    compared to the same baseline (\textsf{base10k}) which includes the blocks. \textsf{Gage} means the generalized-age RvNN, 
    \textsf{Gweight} the generalized-weight RvNN and \textsf{SF} stands for the simple features.}
    \label{fig:noSomethingJustSomething}
\end{figure}

\paragraph{NN Architecture Building Blocks.}

Fig.~\ref{fig:noSomethingJustSomething} shows an analogue of Fig.~\ref{fig:train_and_activations} (left),
plotting the iterative improvement progress for variants of our architecture without one or two of its key building blocks:
the generalized-age RvNN, the generalized-weight RvNN, and the simple features.
We see that combining all three is the best, but each one can be dropped without sacrificing the performance too much.
On the other hand, a single building block alone is always much worse, and the RvNNs alone cannot 
even improve over the baseline after the first iteration of improvement. Interestingly, only using generalized weight
 for the guidance ends up being worse than only using generalized age. Simple features alone are relatively 
 easy to learn and their test performance is actually greater than their train performance here,
 suggesting (not surprisingly) a very low tendency to overfit.

% sudamar2@dai-02:/nfs/sudamar2/lawa$ ./plotter.py ~/mtpa-gnn/newbase10k-cumul ~/mtpa-gnn/newbase10k-cumul-is128 ~/mtpa-gnn/newbase10k-cumul-es48 ~/mtpa-gnn/newbase10k-cumul-es16
\begin{figure}
    \centering
    \includegraphics[scale=0.7]{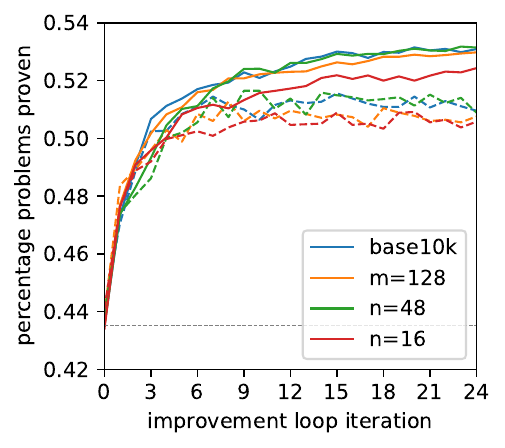}
    \includegraphics[scale=0.7]{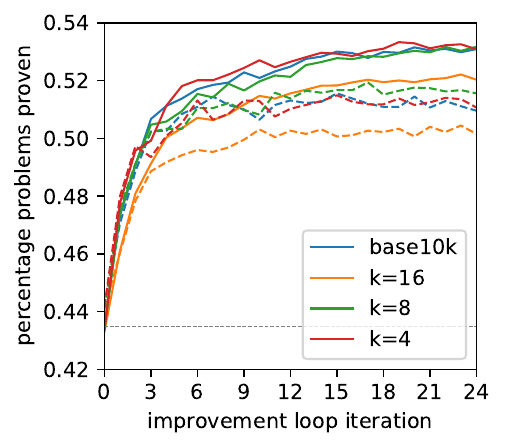}    
    \caption{Performance progress for variants of NN of different embedding size ($n$) and expanded size ($m$) (left),
    and different number of GNN message passing rounds ($k$) (right). Recall that \textsf{base10k} uses $n=32$, $m=256$, and $k=10$. 
    }
    \label{fig:NNsizes}
\end{figure}

\paragraph{NN Size Parameters.}

Fig.~\ref{fig:NNsizes} shows the performance of our architecture when varying the embedding size and the expanded size hyper-parameters (left), and the number of GNN message passing rounds (right). We can see that regarding the first two, at least in the range experimented with,
bigger is always better: Reducing the embedding size to $n=16$ impairs performance noticeably, reducing the expanded size to $m=128$ a bit less, and increasing the embedding size to $n=48$ seems potentially even slightly better than the default in test performance and comparable in train performance. The main reason why we did not use larger values in the main experiment was the increased memory requirement of the training procedure, for which 60 parallel training processes would threaten not to fit into the available \SI{0.5}{\tera\byte} of RAM at peak consumption moments.

On the other hand, the value $k=8$ of GNN message passing round was in our setting at an upper end of the range of favorable values. We see that $k=4$ works comparably well, and $k=16$ is definitely worse. A separate investigation revealed that the latter is mainly not due to the extra time required to process the GNN layers at startup, but rather due to reduced ability to generalize. 
% ./scatter.py ~/mtpa-gnn/newbase10k-cumul-gl8/loop28/train_res.pt ~/mtpa-gnn/newbase10k-cumul-gl16/loop24/train_res.pt
% - it's not that the deeper gnn would be prohibitively slow
% - it's rather that it overfits faster, because its too smart (not enough training data to prevent it from jumping to conclusions)
% This is a nice story to report. Because it shows TPTP is too small for us now!

\begin{table}
    \caption{Model file sizes as function of hyper-parameters $n,m,$ and $k$.}
     \label{tab:filesizes}
    \centering
    \setlength{\tabcolsep}{8pt}
    \begin{tabular}{l|rrr|r}
        session name  & $n$ & $m$ & $k$ &  file size \\
        \midrule
         \textsf{base10k} & 32 & 256 & 10 & \SI{1.9}{\mega\byte} \\
        \midrule
         \textsf{n=48} & 48 & & & \SI{3.6}{\mega\byte}\\
         \textsf{n=16} & 16 & & & \SI{0.8}{\mega\byte} \\
         \textsf{m=128} & & 128 & & \SI{1.7}{\mega\byte} \\
        \midrule
         \textsf{k=4} & & & 4  & \SI{1.0}{\mega\byte} \\
         \textsf{k=8} & & & 8    & \SI{1.6}{\mega\byte} \\
         \textsf{k=16} & & & 16   & \SI{2.8}{\mega\byte} \\
    \end{tabular}

\end{table}

% base10k 1897840
% es48 3649200
% es16 771248
% is128 1717552

% gl4 1008016
% gl8 1601488
% gl16 2787664

The discussed hyper-parameters influence the number of trainable parameters of the network (i.e., $|\bm{\theta}|$), which is reflected by the corresponding file size of the model on the disk. % Note that his also includes to torch script code!
Table~\ref{tab:filesizes} shows the file sizes for the models from Fig.~\ref{fig:NNsizes}.

\end{document}